\documentclass[10pt,twocolumn,letterpaper]{article}

\usepackage[pagenumbers]{cvpr} 
\usepackage{float}
\usepackage{enumitem}
\usepackage[dvipsnames]{xcolor}

\newcommand{\BC}{\boldsymbol{\mathcal{C}}}

\definecolor{commentColor}{RGB}{0,0,0} 
\definecolor{functionColor}{RGB}{0,0,128} 
\usepackage{multirow}

\usepackage{rotating}
\usepackage{graphicx}
\usepackage{multicol}
\usepackage{tabularx}
\usepackage{ulem}
\usepackage{svg}
\usepackage{mwe}
\definecolor{cvprblue}{rgb}{0.21,0.49,0.74}
\usepackage[pagebackref,breaklinks,colorlinks,citecolor=cvprblue]{hyperref}

\def\paperID{15903} 
\def\confName{CVPR}
\def\confYear{2024}

\title{CAD-SIGNet: CAD Language Inference from Point Clouds \\ using Layer-wise Sketch Instance Guided Attention}

\author{
  Mohammad Sadil Khan$^{\dagger}$\\
  {\tt\small mdsadilkhan99@gmail.com}
  \and
  Elona Dupont$^{\dagger}$\\
  {\tt\small elona.dupont@uni.lu}
  \and
  Sk Aziz Ali$^{*}$\\
  {\tt\small sk\_aziz.ali@dfki.de}
  \and
  Kseniya Cherenkova$^{\ddagger\dagger}$\\
  {\tt\small kseniya.cherenkova@uni.lu}
  \and
  Anis Kacem$^{\dagger}$\\
  {\tt\small anis.kacem@uni.lu}
  \and
  Djamila Aouada$^{\dagger}$\\
  {\tt\small djamila.aouada@uni.lu}
  \and
  $^\dagger$SnT, University of Luxembourg, \, $^*$German Research Center for Artificial Intelligence, \, $^\ddagger$Artec3D\\
}

\usepackage{mathtools, stmaryrd}
\DeclarePairedDelimiterX{\Iintv}[1]{\llbracket}{\rrbracket}{\iintvargs{#1}}
\NewDocumentCommand{\iintvargs}{>{\SplitArgument{1}{,}}m}
{\iintvargsaux#1} %
\NewDocumentCommand{\iintvargsaux}{mm} {#1\mkern1.5mu..\mkern1.5mu#2}
\newtheorem{Definition}{Definition}
\renewcommand{\eg}{\textit{e.g.}}

\usepackage{blindtext}
\usepackage{graphicx}
\usepackage[linesnumbered,ruled,vlined]{algorithm2e}

\begin{document}

\twocolumn[{%
\renewcommand\twocolumn[1][]{#1}%
\maketitle
\begin{center}
    \centering
    \captionsetup{type=figure}
    \includegraphics[width=490px,height=100px]{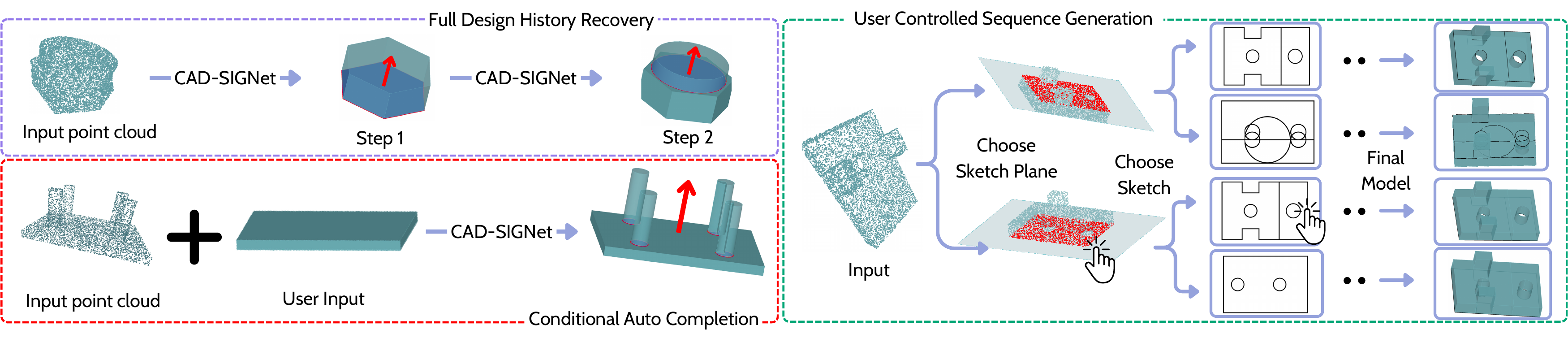}

    \captionof{figure}{Full design history recovery from an input point cloud (top-left) and CAD-SIGNet - user interaction (bottom-left and right).}
       \label{fig:teaser}
\end{center}%
}]


\begin{abstract}
\vspace*{-0.4cm}
Reverse engineering in the realm of Computer-Aided Design (CAD) has been a longstanding aspiration, though not yet entirely realized. Its primary aim is to uncover the CAD process behind a physical object given its 3D scan. We propose CAD-SIGNet, an end-to-end trainable and auto-regressive architecture to recover the design history of a CAD model represented as a sequence of sketch-and-extrusion from an input point cloud. Our model learns CAD visual-language representations by layer-wise cross-attention between point cloud and CAD language embedding. In particular, a new Sketch instance Guided Attention (SGA) module is proposed in order to reconstruct the fine-grained details of the sketches. Thanks to its auto-regressive nature, CAD-SIGNet not only reconstructs a unique full design history of the corresponding CAD model given an input point cloud but also provides multiple plausible design choices. This allows for an interactive reverse engineering scenario by providing designers with multiple next step choices along with the design process. Extensive experiments on publicly available CAD datasets showcase the effectiveness of our approach against existing baseline models in two settings, namely, full design history recovery and conditional auto-completion from point clouds.

\end{abstract}   

\vspace*{-.8\baselineskip}
\section{Introduction}
\label{sec:intro}

    

Computer-Aided Design (CAD) has become the \textit{de facto} method for designing, drafting, and modeling in various industries \cite{robertson,deng}. 
3D reverse engineering is the process of inferring a CAD model given a 3D scan. This procedure requires the expertise of designers and can be time-consuming~\cite{du2018inversecsg,zonegraph}.
Towards the automation of this procedure, several works focused on decomposing point clouds into parametric primitives allowing the reconstruction of the final CAD model~\cite{li2023surface,spfn,sharma2020parsenet,guo2022complexgen,du2018inversecsg}. 
However, CAD modeling 
consists of a sequential process where designers draw 2D sketches (\eg~lines, arcs) and apply CAD operations (\eg~extrusion, chamfer)~\cite{zonegraph,Wu_2021_ICCV}. Recovering these intermediate design steps is crucial as it enables the editablity and re-usability of different object parts sharing the same functionality. 
For instance, a chair can be composed of three design steps, legs, seat, and back rest. Retrieving these steps can allow for editing the legs to be taller, reusing the back rest in another chair design, etc. Nevertheless, identifying adequate design steps requires design expertise. Accordingly, recent methods~\cite{ritchie2023neurosymbolic,Wu_2021_ICCV,willis2021fusion} attempted to learn this expertise from large-scale CAD datasets~\cite{willis2021fusion,Wu_2021_ICCV}. In particular, the sequential nature of CAD modeling made language-like representations with adequate grammar an appealing choice~\cite{CadAsLangNIPS21,xu2022skexgen,Wu_2021_ICCV}. While such a CAD language-like representation has been successfully adopted for CAD generative models~\cite{Wu_2021_ICCV,xu2022skexgen,CadAsLangNIPS21,seff2021vitruvion}, it has not been established for 3D reverse engineering. As in point cloud captioning~\cite{cai20223djcg,chen2021scan2cap}, leveraging language-like representations for reverse engineering requires mechanisms for jointly learning visual representation from point clouds and corresponding CAD language. Hence, the main question that we ask is: \textbf{\textit{how to effectively learn CAD visual-language representations from point cloud and CAD sequences for 3D reverse engineering?} }











\begin{figure*}[!ht]
\centering
    \includegraphics[width=.89\linewidth]{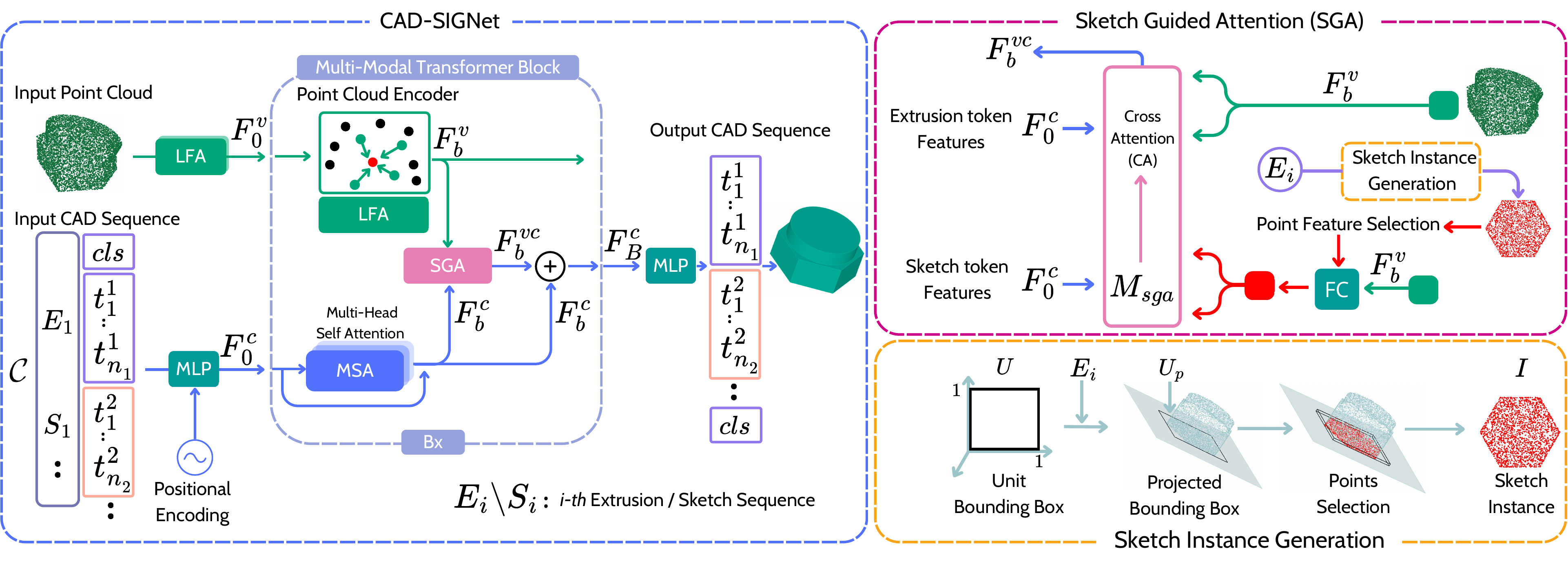}
    \vspace*{-.8\baselineskip}
    \caption{\textbf{Method Overview}. CAD-SIGNet (left) is composed of $\mathbf B$ Multi-Modal Transformer blocks, each consisting of an $\operatorname{LFA}$~\cite{randlanet} module to extract point features, $\mathbf F_{b}^v$, and a MSA~\cite{transformer} module for token features, $\mathbf F_{b}^c$. A SGA module (top right) combines $\mathbf F_{b}^v$ and $\mathbf F_{b}^c$ for CAD visual-language learning. A sketch instance (bottom right), $\mathbf I$, obtained from the predicted extrusion tokens is used to apply a mask, $\mathbf M_{\text{sga}}$ during CA to predict sketch tokens.}
    \label{fig:architecture}
  \vspace*{-.8\baselineskip}
\end{figure*}

\vspace{0.05cm}
To answer this question, many challenges need to be addressed due to the structural disparity between point clouds in 3D space and language-like representations of CAD sequences~\cite{multicad}. In particular, CAD sequences encode both the chronological order of design steps and their parametric form~\cite{Wu_2021_ICCV,xu2022skexgen}, while the corresponding point clouds only encode the geometry of the final design~\cite{du2018inversecsg}. 
To the best of our knowledge, the only works that infer CAD language from point clouds are DeepCAD~\cite{Wu_2021_ICCV} and MultiCAD~\cite{multicad}. While DeepCAD~\cite{Wu_2021_ICCV} focused on learning CAD language using a feed-forward strategy and presented the point cloud to CAD language setting as a future application, MultiCAD~\cite{multicad} focused on learning the interaction of features from distinct modalities (\ie~point cloud and CAD language) through a contrastive learning framework. Despite their promising results, both methods suffer from two main limitations: (1)~Both visual and CAD language representations are learned separately in the first stage. A mapping between the two representations is learned afterwards. Nevertheless, this separate learning might result in modality-specific features that are not relevant for CAD language inference from point clouds~\cite{sun2019videobert}; (2)~the learning of CAD language representation is achieved using a feed-forward strategy where the CAD language of the full design history is inferred at once. However, in a real-world scenario, providing input or preferences at each design steps would allow for tailoring the solution to the requirements of the designer~\cite{hnc,xu2022skexgen}. 
\vspace*{-.3\baselineskip}
To address the aforementioned challenges and limitations, we propose \textit{CAD-SIGNet}, an end-to-end trainable architecture that auto-regressively infers CAD language in the form of \textit{sketch-and-extrusion} design steps from point clouds. Instead of learning separate representations for both point clouds and CAD language and the mapping between them, the proposed method jointly learns these representations through multi-modal transformer blocks. Each block is composed of layer-wise cross-attention between CAD language and point cloud embedding. 
Moreover, other existing works~\cite{Wu_2021_ICCV,multicad} infer sketches from a global representation of the point cloud. However, we assume that only a subset of the point cloud is needed to parameterize a sketch. As shown in the right panel of Figure~\ref{fig:teaser}, designers specify a plane in 3D space where the sketch is drawn. The intersection of the sketch region and the point cloud (shown in red in the same Figure) is assumed to be sufficient for sketch parameterization. Therefore, this subset, referred to as \textit{Sketch Instance}, is first identified and then considered in the cross-attention to infer sketch parameters. We refer to this technique as \textit{Sketch instance Guided Attention} (SGA). It allows the network to focus its attention on specific points (\ie sketch instance), hence improving fine-grained sketch inference. Finally, the auto-regressive nature of CAD-SIGNet allows multiple plausible design choices to coexist. As shown in the right panel of Figure~\ref{fig:teaser}, this enables an interactive reverse engineering scenario, offering designers various choices throughout the CAD process. An overview of the proposed approach is provided in Figure~\ref{fig:architecture}.







\noindent \textbf{Contributions:} The main contributions can be summarized as follows: 

\begin{itemize}
    \item An end-to-end trainable auto-regressive network that infers CAD language given an input point cloud. 
    To the best of our knowledge, we are the first to propose an auto-regressive strategy for this problem. 
    \item Multi-modal transformer blocks with a mechanism of layer-wise cross-attention between point cloud and CAD language embedding.
    \item A Sketch instance Guided Attention (SGA) module which guides the layer-wise cross-attention mechanism to attend on relevant regions of the point cloud for predicting sketch parameters.  
    \item A thorough experimental validation in two different reverse engineering settings, namely, full CAD history recovery and conditional auto-completion from point clouds (see bottom left panel of Figure~\ref{fig:teaser}).
\end{itemize}

  \vspace*{-.8\baselineskip}


\      
\vspace{-.9\baselineskip}
\section{Related Works }
\label{sec:related_work}
\noindent{\textbf{Deep Learning-based CAD Reverse Engineering:}} CAD models are well defined 3D objects described by their geometric and topological properties. As such, some works address the reverse engineering problem by focusing on recovering the geometric features of CAD models from point clouds. This has been achieved using parametric fitting techniques either on the edges of the CAD model~\cite{wang2020pie, liu2021pc2wf, Cherenkova_2023_CVPR, mallis2023sharp} or on the surfaces~\cite{fischler1981random, spfn,sharma2020parsenet, le2021cpfn, yan2021hpnet,guo2022complexgen}. 
However, a parametric fitting approach can only provide information about the final CAD model and it lacks any insight into the design process and the intermediate steps that were used to create the CAD model.
In order to address these limitations, another line of work~\cite{Sharma_2018_CVPR, du2018inversecsg, friedrich2019optimizing, kania2020ucsg, Yu_2022_CVPR, yu2023d} models the CAD construction using \textit{Constructive Solid Geometry}~(CSG)~\cite{kania2020ucsg}. CSG is a sequential method in CAD modeling that combines simple 3D shapes (e.g., cube, sphere) using boolean operations (e.g., union, intersection). While CSG can allow for the construction of relatively complex shapes, it is no longer the standard in the CAD industry~\cite{zonegraph}. Indeed, the \textit{feature-based} approach has now been adopted by most CAD software as it allows for the modelling of more complex shapes using a sequence of sketches and CAD operations~\cite{willis2021fusion}. 
The work in~\cite{uy2022point2cyl} attempts to retrieve some of the features of the construction history as extrusion cylinders, but requires manual input to combine the cylinders into the final shape and does not result into parametric sketches. Self-supervised~\cite{li2023secad} and unsupervised~\cite{ren2022extrude} approaches have also been adopted in this context. Nevertheless, these approaches strive to infer plausible design steps approximating the input point cloud, but not necessarily inferring the standard parametric entities and therefore not reproducing design expertise. CAD-SIGNet goes beyond these limitations and leverages feature-based sequences of real design steps to predict CAD history from point clouds. \\
\noindent{\textbf{CAD as a Language:}} Due to the sequential nature of feature-based CAD modeling, a common strategy to represent it is to use language modelling. Inspired by Natural Language Processing (NLP)~\cite{transformer}, some works have focused on language modeling of CAD sketches~\cite{CadAsLangNIPS21, seff2021vitruvion,li2022free2cad}, others leveraged it in the context of CAD models~\cite{hnc, xu2022skexgen}. However, all the aforementioned works present generative models that allow for the manipulation of a latent space but do not directly tackle the reverse engineering problem. CADParser~\cite{zhoucadparser2023} used an intermediate representation of the final shape, called \textit{Boundary-Representation} (B-Rep)~\cite{lambourne2021brepnet}, instead of point cloud to relax the problem of CAD language inference. Closest to our work are DeepCAD~\cite{Wu_2021_ICCV} and MultiCAD~\cite{multicad}. DeepCAD proposed a language-based sketch-extrusion formulation and predicted the CAD history from point clouds as a preliminary experiment.
Building on these findings, MultiCAD~\cite{multicad} opted for a two-stage multimodal contrastive learning strategy. In addition to the separate modality learning, both~\cite{Wu_2021_ICCV} and~\cite{multicad} use a feed-forward strategy limiting the scope of reverse engineering scenarios. In contrast, CAD-SIGNet presents a joint visual-language learning strategy and allows designers to interact with design choices (see Figure~\ref{fig:teaser}).
\section{Problem and CAD Language Formulation}
\label{sec:input_representation}

Given an input point cloud, our objective is to generate a sequence of tokens representing the design history of the corresponding CAD model. 
Formally, let $\mathbf X=\lbrack~\mathbf x_1,\mathbf x_2,\dots,\mathbf x_N\rbrack\in\mathbb{R}^{N\times 3}$ be an input point cloud with $\mathbf x_i\in\mathbb{R}^{3}$ denoting the 3D coordinates of the $i$-th point and $N$ the number of points. Following recent CAD generative models~\cite{Wu_2021_ICCV,xu2022skexgen}, the design history of a CAD model $\BC=\{\BC_j\}_{j=1}^{n_s}$ is represented by a sequence of $n_s$ design steps, where each step $\BC_j=~\{t_k\}_{k=1}^{n_j}$ consists of a sequence of $n_j$ tokens $t_k\in\Iintv{0,d_t}$, with $d_t$ defining the tokenization interval. The objective is to learn a mapping, 
\vspace*{-.3\baselineskip}
\begin{align*}
\mathbf{\Phi} : \mathbb{R}^{N\times 3 }  \rightarrow{\Iintv{0,d_{t}}}&^{n_{ts}} \,\,\,\, \text{s.t,} \,\,\, \mathbf{\Phi}(\mathbf{X})=  \BC
\end{align*}
\noindent where $n_{ts} = \sum_{j=1}^{n_s} n_j$ denotes the total number of tokens. As in~\cite{Wu_2021_ICCV,xu2022skexgen}, the design history is assumed to be composed of \textit{sketch-and-extrusion} sequences. This implies that the sequence of tokens $\{t_k\}_{k=1}^{n_j}$ of each design step $\BC_j$ represents either a parametric sketch $\mathbf S$ or an extrusion operation $\mathbf E$ and the full design history $\BC$ can be seen as a sequence of sketch-and-extrusion pairs $\{(\mathbf S_l,\mathbf E_l)\}_{l=1}^{n_s/2}$. 

\begin{figure}
\centering
   \includegraphics[width=.9\linewidth]{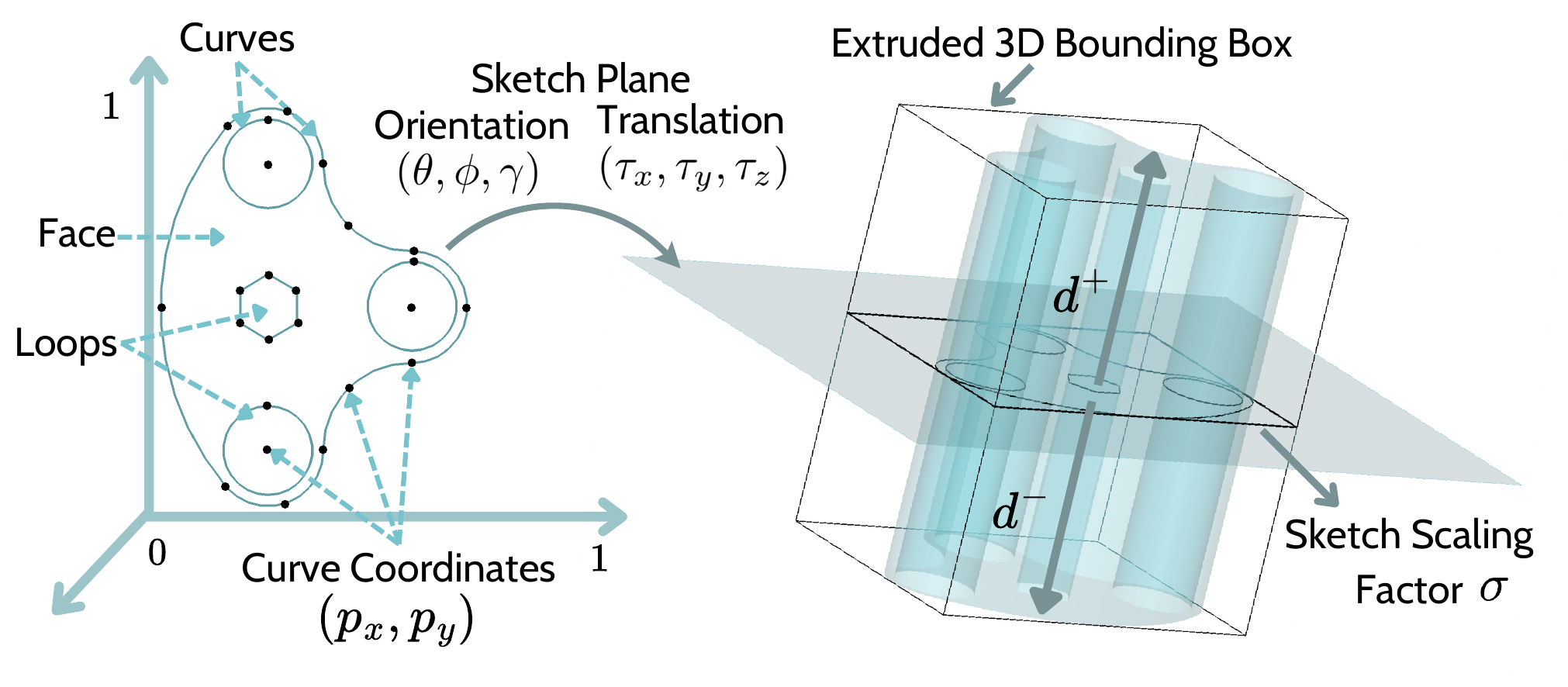}
   \vspace*{-.6\baselineskip}
\caption{Illustration of sketch and extrusion representations.}
\label{fig:sketch_extrude_representation}
\vspace*{-.9\baselineskip}
\end{figure}

\vspace{0.05cm}
\noindent \textbf{Sketch Representation:} Similarly to~\cite{xu2022skexgen}, a hierarchical representation of the sketch is considered. As depicted in Figure~\ref{fig:sketch_extrude_representation}, a sketch is created from one or more faces, with a face being a 2D region bounded by loops. A loop, in turn, is a closed path that can consist of either a single closed curve, such as a circle, or multiple curves, e.g., combination of lines and arcs. The curves are represented by the tokenized 2D coordinates $(p_x,p_y)$ of their parametric formulation (e.g., start and end points for lines). The end of a curve, loop, face, and sketch are represented by the end tokens $e_c$, $e_l$, $e_f$, and $e_s$, respectively.  

\noindent \textbf{Extrusion Representation:} The extrusion operation defines the sketch plane and the parameters needed to turn it into a 3D volume. Following~\cite{xu2022skexgen}, the tokens $(\theta, \phi, \gamma)$ and $(\tau_x, \tau_y, \tau_z)$ define the sketch plane orientation and translation, respectively, with respect to a reference coordinate system. The token $\sigma$ scales the normalized sketch defined by the sketch tokens. The pair $(d^+, d^-)$ represents the extrusion distances along the normal direction of the sketch plane and its opposite, respectively. The parameter $\beta$ denotes the type of extrusion operation among \textit{new}, \textit{cut}, \textit{join}, and \textit{intersect}. Finally, 
$e_e$ sets the end of the extrusion tokens. Figure~\ref{fig:sketch_extrude_representation} shows the different tokens used to represent the extrusion operation. \\
In addition to sketch and extrusion tokens, $pad$ is used for padding and $cls$ is considered to indicate the start or end of the design sequence. More details about CAD sequence representation are provided in supplementary materials.

\vspace{-.2\baselineskip}
\section{CAD-SIGNet Architecture}
\label{sec:CAD-SIGNet}
The proposed CAD-SIGNet is an end-to-end trainable transformer-based architecture that takes a point cloud $\mathbf X$ as an input and outputs the corresponding design history sequence $\BC$. It follows an auto-regressive strategy by considering the set of previous tokens $\BC_{<i}=\{t_j\}_{j<i}$ as context to infer the next token $t_i$. For a given point cloud $\mathbf X$, the goal of CAD-SIGNet is to learn its corresponding CAD history using the following probability distribution,
\vspace*{-.6\baselineskip}
\begin{equation}
p_{\theta}(\BC \mid \mathbf X) = \prod_{i=1}^{n_{ts}} p_{\theta}(t_i \mid \{t_j\}_{j<i}, \mathbf{X}) \ ,
\label{eq:autoreg}
\vspace*{-.6\baselineskip}
\end{equation}
\noindent where $t_i$ is the $i$-th sequence token and $\theta$ denotes the learned parameters of the network. As mentioned in Section~\ref{sec:input_representation}, the predicted tokens $t_i$ correspond to the representations of sketch-and-extrusion sequences. Unlike other CAD language generative models~\cite{xu2022skexgen,Wu_2021_ICCV} which infer sketch tokens $\mathbf S_k$ for each design step $\BC_k$ followed by extrusion tokens $\mathbf E_{k+1}$, CAD-SIGNet first predicts extrusion tokens that are further used as context to predict sketch tokens. An overview of our CAD-SIGNet modules is provided in the left panel of Figure~\ref{fig:architecture}. 


\vspace*{-.3\baselineskip}
\subsection{Point Cloud and CAD Language Embedding}
The first module of CAD-SIGNet is responsible for embedding point cloud points and CAD language tokens into the same $d_e$-dimensional space $\mathbb R^{d_e}$.


\noindent \textbf{Point Cloud Embedding:} Given the point cloud $\mathbf X\in\mathbb{R}^{N\times(3+f)}$, where $f$ is the number of additional per-point estimated features\footnote{Point normals are extracted using Open3D~\cite{open3d}}, a linear layer\footnote{All linear layers used in the paper consist of a weight matrix and a bias. For notation simplicity, we omit the bias. } followed by ReLU~\cite{relu} is firstly applied as follows, 
\vspace*{-.3\baselineskip}
\begin{equation}\label{eq:pc_emb}
   \mathbf F_0^p = \operatorname{ReLU}(\mathbf X\mathbf W_{\text{emb}}^p) \ ,
\end{equation}
\noindent where $\mathbf F_0^p\in\mathbb{R}^{N\times d_e^{p_0}}$ is the learned embedding, $\mathbf W_{\text{emb}}^p\in\mathbb{R}^{(3+f)\times d_e^{p_0}}$ is a learnable matrix, and $d_e^{p_0} = 16$. The per-point features obtained in $\mathbf F_0^p$ are further enriched using two~\textit{Local Feature Aggregation} (LFA)~\cite{randlanet} modules. LFA uses k-Nearest Neighbor (k-NN) to aggregate the features of neighboring points through a linear combination weighted by learned attention weights. A linear layer is applied on the resulting aggregated features followed by ReLU for each LFA module. The first LFA module results in the point cloud embedding $\mathbf F_0^v\in\mathbb R^{N\times d_e}$ defined by,
\vspace*{-.3\baselineskip}
\begin{equation}
\begin{aligned}
    \mathbf F_0^v = \operatorname{ReLU}(\operatorname{LFA}(\mathbf F_0^p)\mathbf W_{\text{lfa}}) \ , 
    \label{eq:lfa-pc}
\end{aligned}
\vspace*{-.3\baselineskip}
\end{equation}
\noindent where $\mathbf W_{\text{lfa}}\in\mathbb R^{d_e^{p_0} \times d_e}$ denotes the weight matrix of the linear projection. The second $\operatorname{LFA}$ module is applied on $ \mathbf F_0^v$ without changing its dimension. For more details about the operator $\operatorname{LFA}(.)$, readers are referred to~\cite{randlanet}. 

\vspace{0.05cm}
\noindent \textbf{CAD Language Embedding:} 
Given an input design sequence $\BC=\{t_i\}_{i=1}^{n_{ts}}\in\Iintv{0,d_{t}}^{n_{ts}}$, a matrix form of the sequence is adopted. Unlike~\cite{xu2022skexgen} which maps the sketch coordinates $p_x$ and $p_y$ into a \hbox{$1$-dimensional} token, we consider them as a single \hbox{$2$-dimensional} token $(p_x,p_y)$. To avoid dimension mismatch, the other tokens are also considered as $2$-dimensional by augmenting them with $pad$ tokens. By concatenating these tokens and using a one-hot encoding, a matrix form $\mathbf C\in\{0,1\}^{n_{ts}\times2d_t}$ is used to represent the sequence $\BC$. 
As in~\cite{xu2022skexgen}, token flags $\mathbf C_{\text{type}}\in\Iintv{0,n_f}^{n_{ts}\times1}$ and $\mathbf C_{\text{step}}\in\Iintv{0,n_s/2}^{n_{ts}\times1}$ are set to indicate token types and design steps, respectively. The initial embedding of the CAD language $\mathbf F_0^c\in\mathbb{R}^{n_{ts}\times d_e}$ is obtained by using the aforementioned token representations within a linear layer and is given by,
\vspace*{-.3\baselineskip}
\begin{equation}
    \mathbf F_0^c=[\mathbf C+\mathbf M_{\text{seq}},\mathbf C_{\text{type}},\mathbf C_{\text{step}}]\mathbf W_{\text{emb}}^c+\mathbf C_{\text{pos}} \ ,
    \label{eq:cad_emb}
    \vspace*{-.4\baselineskip}
\end{equation}

\noindent where $(,)$ is the concatenation operation, $\mathbf W_{\text{emb}}^c\in\mathbb{R}^{(2d_t+2)\times d_e}$ is a learnable weight matrix, and $\mathbf C_{\text{pos}}\in\mathbb R^{n_{ts}\times d_e}$ a learned positional encoding. Note that CAD sequences have a variable number of tokens $\Tilde{n}_{ts}<n_{ts}$ and $\mathbf M_{\text{seq}}\in\{0,-\infty\}^{n_{ts}\times 2d_t}$ is the padding mask that sets token embedding beyond $\Tilde{n}_{ts}$ to~$-\infty$.







\vspace*{-.3\baselineskip}
\subsection{Layer-wise Multi-Modal Transformer Block}
\label{sec:cross-attention}
Based on the aforementioned embedding, CAD-SIGNet jointly learns visual-language representations using $B$ multi-modal transformer blocks of layer-wise cross-attention between CAD and point cloud embedding.
\vspace{0.1cm}

In particular, let the CAD language embedding $\mathbf F_{b-1}^c$ and the point cloud embedding $\mathbf F_{b-1}^v$ be the input of the $b$-th block (i.e., the first block receives $\mathbf F_0^c$ defined in Eq.~(\ref{eq:cad_emb}) and $\mathbf F_{0}^v$ defined in Eq.~(\ref{eq:lfa-pc})).
Firstly, $\mathbf F_{b}^c$ is generated from $\mathbf F_{b-1}^c$ using a multi-head scaled dot-product attention~\cite{transformer}~($\operatorname{SA}$) and an add-normalization layer~\cite{transformer}~($\operatorname{AddNorm}$) as follows
\vspace*{-.9\baselineskip}
\begin{align}
    \mathbf{F}_{b}^c &= \operatorname{SA}(\mathbf{Q},\mathbf{K},\mathbf{V},\mathbf{M})\label{eq:scaled-dot-product}\\
    \mathbf{F}_{b}^{c} &= \operatorname{AddNorm}(\mathbf{F}_{b}^{c},\mathbf{F}_{b-1}^{c})
    \label{eq:add_norm}
\end{align} \noindent where query $\mathbf Q$, key $\mathbf K$, and value $\mathbf V$ are extracted from $\mathbf F_{b-1}^c$ and $\mathbf M$ is the standard self-attention mask~\cite{transformer}.
On the other hand, the point cloud embedding $\mathbf F_{b-1}^v$ undergoes an additional LFA module as described in Eq.~(\ref{eq:lfa-pc}) to obtain a point cloud embedding $\mathbf F_{b}^v~\in~\mathbb R^{N \times d_e}$. 
\vspace{0.1cm}

To enable the information passing between CAD language and point cloud embedding within each block, a cross-attention layer is used on $\mathbf F_b^v$ and $\mathbf F_{b}^{c}$. This is achieved by employing linear projections to extract a key $\mathbf K_v\in\mathbb R^{N \times d_e}$ and value $\mathbf V_v\in\mathbb R^{N \times d_e}$ from the point cloud embedding $\mathbf F_b^v$. A query $\mathbf Q_c\in\mathbb R^{n_{ts} \times d_e}$ is extracted from the CAD embedding $\mathbf F_{b}^{c}$. 
Using Eq.~\eqref{eq:scaled-dot-product}, the cross-attention layer computes a CAD visual-language embedding $\mathbf F_{b}^{vc}~\in~\mathbb R^{n_{ts}\times d_e}$ as follows,
\vspace*{-.3\baselineskip}
\begin{equation}
    \mathbf F_{b}^{vc} = \operatorname{SA}(\mathbf Q_c,\mathbf K_v,\mathbf V_v,\mathbf 0) \ ,
    \label{eq:cross-attention}
\end{equation}
\vspace*{-.04\baselineskip}
\noindent where $\mathbf 0$ is a $({n_{ts}\times N})$ zero matrix. Furthermore, the cross and self-attended embedding $\mathbf F_{b}^{vc}$ and $\mathbf F_{b}^{c}$ are added and normalized to help the network learn the geometric relationship between CAD tokens, yielding, $ \mathbf F_{b}^{c}~=~\operatorname{AddNorm}(\mathbf F_{b}^{vc},\mathbf F_{b}^{c})$.
Finally, as in~\cite{transformer}, a Feed-Forward Network (FFN) is applied on $\mathbf F_{b}^{c}$ and added to it to form the final CAD embedding, which is passed to the next block along with $\mathbf F_{b}^{v}$. 
\begin{figure*}[h]
\centering
    \includegraphics[width=0.95\linewidth]{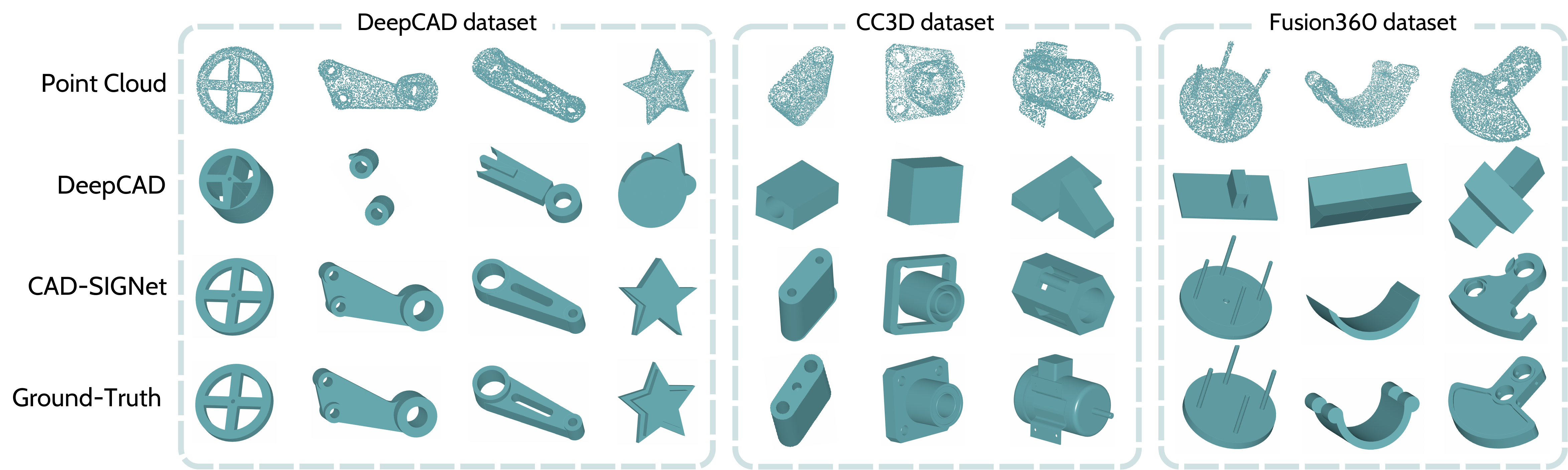}
    \caption{Visual results of reconstruction from the CAD sequences predicted from input point clouds. Both DeepCAD~\cite{Wu_2021_ICCV} and CAD-SIGNet are trained on DeepCAD dataset~\cite{Wu_2021_ICCV}. \textbf{Left}: Results on DeepCAD dataset~\cite{Wu_2021_ICCV}. \textbf{Middle}: Cross-dataset results on CC3D dataset~\cite{cc3d}, \textbf{Right}: Cross-dataset results on Fusion360 dataset~\cite{willis2021fusion}.}
    \label{fig:vis-res}
      \vspace*{-.7\baselineskip}
\end{figure*}

\vspace{0.05cm}
\noindent\textbf{Sketch Instance Guided Attention (SGA):} The aforementioned multi-modal transformer blocks are designed to pass the information from all point embedding to CAD token embedding. 
However, we posit that parameterizing a sketch requires only cross-attending to a subset of the point cloud. As shown in the bottom right panel of Figure~\ref{fig:architecture}, the intersection between the sketch region and the point cloud (depicted in red in the same Figure) is considered as adequate for sketch parameterization. 
As depicted in Figure~\ref{fig:sketch_extrude_representation}, the representation of extrusion tokens defines the sketch plane and bounding box. Furthermore, CAD-SIGNet predicts extrusion tokens followed by sketch tokens for each design step. This implies that the predicted extrusion tokens can be leveraged to define a \textit{sketch instance} on the point cloud for cross-attention with sketch token embedding. \vspace*{-.3\baselineskip}
\indent \begin{Definition} A sketch instance $\mathbf I\in\mathbb{R}^{\eta \times 3}\subset\mathbf X$, with $\eta<N$, is a subset of the input point cloud $\mathbf X$. It is extracted by selecting points inside the bounding box on the sketch plane derived from the corresponding predicted extrusion tokens.  
\end{Definition}
\vspace*{-.1\baselineskip}
The bottom right panel of Figure \ref{fig:architecture} shows the sketch instance extraction process. Given a set of extrusion tokens $\mathbf E$, we first project the unit bounding box of the $xy$-plane\footnote{Sketches are normalized to fit a unit bounding box in the $xy$-plane.} into a bounding box on the sketch plane defined by the extrusion tokens $\mathbf E$. In particular, given the unit bounding on $xy$-plane defined by the points $\mathbf U~=~[(0,0,0)^{\mathbf T},(0,1,0)^{\mathbf T},(1,0,0)^{\mathbf T}]~\in~\mathbb R^{3\times3}$, the Euler angles $(\theta,\phi,\gamma)$, the translation vector $(\tau_x,\tau_y,\tau_z)$, and the scaling factor $\sigma$ defined by the extrusion operation $\mathbf E$, the projected bounding box $\mathbf U_p~\in~\mathbb R^{3\times3}$ is given by,
\vspace*{-.3\baselineskip}
\begin{equation}
    \mathbf U_p = (\mathbf R_{xyz}(\theta,\phi,\gamma)(\mathbf U  \times \sigma) + \mathbf (\tau_x,\tau_y,\tau_z)^{{\mathbf T}}) \ ,
\end{equation}
\noindent where $\mathbf R_{xyz}(\theta,\phi,\gamma)\in\mathcal{SO}(3)$ combines the Euler angles in a rotation matrix in the special orthogonal group $\mathcal{SO}(3)$. The sketch instance $\mathbf I$ is then defined by the points of $\mathbf X$ lying inside this bounding box, i.e., $\mathbf I:=\{\mathbf x\in\mathbf X|\phi(\mathbf x,\mathbf U_p)=\text{True}\}$, where $\phi(\mathbf x,\mathbf U_p)$ is an operator that checks whether an input point $\mathbf x\in\mathbb R^3$ is inside the projected bounding box $\mathbf U_p$. 
Note that for training the ground-truth extrusion tokens are used to define the bounding box $\mathbf U_p$, while the predicted extrusion tokens are leveraged at inference time. In order to not penalize small errors in sketch plane predictions of the extrusion tokens and point cloud sampling, the bounding box is enlarged in the direction of sketch plane normal and its opposite by a small margin $0.1\times\operatorname{max}(d^+,d^-)$. 
The extracted sketch instances can be then used in the cross-attention defined in Eq.~({\ref{eq:cross-attention}) only for sketch token embedding by employing a suitable mask instead of the zero matrix. In particular, let $\mathbf M_{\text{sga}}\in\{0,-\infty\}^{n_{ts}\times N}$ be this mask and $m^{\text{sga}}_{ij}$ be its value for the attention between the $i$-th token and $j$-th point embedding. $\mathbf M_{\text{sga}}$ is introduced to mask the attention of sketch token embedding to the points lying outside their corresponding sketch instance. As a result, $m^{\text{sga}}_{ij}$ is set to $0$ if the $i$-th token embedding is not denoting a sketch. If the $i$-th token is representing a sketch, then $m^{\text{sga}}_{ij}$ is set to $0$ where the $j$-th point embedding is part of the corresponding sketch instance and $-\infty$ otherwise. Note that after identifying the sketch instances, $4$ linear layers are used on the corresponding subsets of $\mathbf F_b^v$ to refine their embedding before extracting the key and value for the cross-attention with sketch token embedding. The top right panel of Figure~\ref{fig:architecture} visually describes the SGA module. 

\vspace*{-.3\baselineskip}
\subsection{Training and Inference Strategies}
\vspace*{-.3\baselineskip}
\label{sec:train-inference}
After the last multi-modal transformer block, the CAD embedding $\mathbf F_B^c$ is passed to two separate linear layers for predicting the 2D tokens probability matrices $\mathbf O_x,~\mathbf O_y~\in~[0,1]^{n_{ts}\times d_t}$. 

\vspace{0.05cm}
\noindent \textbf{Training:} During training, a teacher-forcing strategy~\cite{teacher-forcing} is used to pass the ground-truth as input. The cross-entropy loss $\boldsymbol{\mathcal{L}}_{ce}$ is used as an objective function.

\vspace{0.05cm}
\noindent \textbf{Inference:} During inference, given the input point cloud $\mathbf X$ and the initial CAD sequence consisting of $\BC~=~\{(cls,pad)\}$, the next tokens are auto-regressively generated until the $\textit{end}$ token is predicted.

\vspace{0.05cm}
\noindent \textbf{Hybrid Sampling:} The auto-regressive nature of CAD-SIGNet suggests that different token predictions at a given time-stamp result in different final CAD sequences. 
This allows for generating multiple plausible predictions given a point cloud. In particular, given the output probabilities $\mathbf O_x, \mathbf O_y$, one can take the \textit{top-$1$} to obtain the predicted tokens or opt for a different selection strategy for each token to have a different final CAD sequence. To showcase this, we use a hybrid sampling approach during inference by selecting \textit{top-$5$} probabilities for the first token, and \textit{top-$1$} for subsequent tokens. This results in $5$ different final CAD sequences given a point cloud. 
Finally, the optimal CAD sequence is chosen by selecting the one that best approximates the input point cloud. This is assessed by reconstructing the CAD models\footnote{Opencascade\cite{opencascade} is used to reconstruct a model from a CAD sequence.} from the predicted sequences, sampling point clouds on them, and selecting the model that results in a minimum Chamfer Distance~\cite{chamfer_distance} with respect to the input point cloud. 

\vspace{-.2\baselineskip}
\section{Experimental Results} 
\label{sec:experiment}
\vspace{-.3\baselineskip}

In this section, the proposed CAD-SIGNet is evaluated on two reverse engineering scenarios: (1) design history recovery from point clouds, and (2) conditional auto-completion of design history given user input and point clouds. Additional preliminary experiments showcasing the applicability of the proposed method in a realistic scenario of reverse engineering is also discussed.

\vspace{0.05cm}
\noindent \textbf{Dataset}: The DeepCAD dataset \cite{Wu_2021_ICCV} is used.
The sketch and extrusion parameters are normalized, ensuring that the sketches and the CAD models are within a unit-bounding box starting from the origin. The point clouds are obtained by uniformly sampling $8192$ points from the normalized CAD model. As in~\cite{Wu_2021_ICCV}, the sketch and extrusion parameters are quantized to $8$ bits. 

\vspace{0.05cm}
\noindent \textbf{Implementation Details}: 
We use $8$ CAD-SIGNet multi-modal transformer blocks with $h=8$ number of heads for self-attention. The latent dimension is set to $d_e=128$. 
The network has been trained 
with a batch size of $72$ for $150$ epochs using $2$ \textit{NVIDIA RTX A6000} GPUs. 
We implement curriculum learning~\cite{curriculum-learning} for the first $15$ epochs, increasingly sorting CAD sequences by the number of curves. For the subsequent $135$ epochs, the samples are shuffled. More details are provided in supplementary materials. 
\vspace*{-.14\baselineskip}
\subsection{Design History Recovery from Point Cloud} \label{sec:design-history}
\vspace*{-.1\baselineskip}
In this experiment, the task is to infer CAD language history given an input point cloud. 


\vspace{0.05cm}
\noindent \textbf{Metrics}: To thoroughly evaluate the predicted sequences, a set of metrics assessing different levels of the predictions are used. In particular, the final CAD reconstructions are quantitatively evaluated with respect to ground-truth CAD models using mean and median \textit{Chamfer Distances} (CD)~\cite{chamfer_distance}. As CAD sequences are predicted as tokens, they do not necessarily result in valid CAD models when reconstructed using OpenCascade~\cite{opencascade}. Accordingly, an \textit{Invalidity Ratio} (IR) metric, expressed as a percentage, is the ratio of invalid models. 
%
\textit{F1} score is computed to evaluate the predicted extrusions and different primitive types along with their occurrences in the sequences.
A Hungarian matching algorithm~\cite{hungarian-matching} is used to match the predicted loop and primitive bounding boxes with the ground-truth ones of the same sketch. More details on the metrics are provided in supplementary materials. 
\begin{table}[h]
\vspace*{-.5\baselineskip}
\resizebox{\linewidth}{!}{
    \begin{tabular}{lccc}
\hline
Model              & \begin{tabular}[c]{@{}c@{}}IR$\downarrow$\\\end{tabular} & \begin{tabular}[c]{@{}c@{}}Mean CD $\downarrow$\\$(\times 10^3)$\end{tabular}& \begin{tabular}[c]{@{}c@{}}Median CD $\downarrow$\\$(\times 10^3)$\end{tabular}\\
\hline
 DeepCAD~\cite{Wu_2021_ICCV}& 7.14& 42.49&9.640\\
 MultiCAD~\cite{multicad}& 11.5& -&8.090\\
\textbf{CAD-SIGNet (Ours)}                                   & \textbf{0.88}&                                                                \textbf{3.430}&                                                                   \textbf{0.283}\\ \hline
\end{tabular}}
\vspace*{-.5\baselineskip}
    \caption{Design history recovery from point clouds on DeepCAD~\cite{Wu_2021_ICCV} dataset. Invalidity Ratio (IR), mean and median Chamfer Distance (CD) results.}
    \label{tab:icd}  
    \vspace*{-.8\baselineskip}
\end{table}

\noindent \textbf{Results:} 
To the best of our knowledge, DeepCAD and MultiCAD are the only works in literature that perform point cloud to CAD language inference. Note that DeepCAD results have been reproduced using its publicly available implementation, while MultiCAD results were taken from the reported ones in~\cite{multicad} due to unavailability of public implementation. It can be observed in Table~\ref{tab:icd} that the proposed CAD-SIGNet is outperforming both DeepCAD and MultiCAD in terms of median CD by a factor of $35$ and $28$, respectively. Moreover, the IR metric shows that the predicted CAD sequences by CAD-SIGNet results in drastically more valid CAD model reconstructions than both DeepCAD and MultiCAD. In Table~\ref{tab:primitive}, the per-primitive and extrusion F1 scores of our method are compared to those of DeepCAD. Our model predicts more accurately the primitive types, and their occurrences in the design sequence when compared to DeepCAD. Notably, our method records improvements of more than $14\%$ in F1 score on the arc type with respect to DeepCAD. In addition, CAD-SIGNet correctly predicts the extrusions in most cases, showing that our model can correctly identify the number of needed design steps. Figure~\ref{fig:vis-res} displays several qualitative CAD models reconstructed from the predicted CAD sequences. Visually, our method achieves better reconstructions with more accurate details than DeepCAD~\cite{Wu_2021_ICCV}. More visual results 
are reported in the supplementary materials. 

\begin{table}[!h]
\vspace*{-.5\baselineskip}
\resizebox{\linewidth}{!}{
\begin{tabular}{lcccc}
\hline
\multirow{2}{*}{Model} & \multicolumn{4}{c}{F1 Score}                                                           \\
                       & Line $\uparrow$ & Arc$\uparrow$& Cricle$\uparrow$& \multicolumn{1}{c}{Extrusion$\uparrow$} \\ \hline
DeepCAD\cite{Wu_2021_ICCV}               & 69.26          & 13.82          & 60.14          & 86.70                         \\
\textbf{CAD-SIGNet (Ours)}      & \textbf{77.31} & \textbf{28.65} & \textbf{70.36} & \textbf{92.72}                \\ \hline
\end{tabular}
}
\vspace*{-.5\baselineskip}
\caption{Design history recovery from point clouds on DeepCAD~\cite{Wu_2021_ICCV} dataset. F1 scores for lines, arcs, circles, and extrusions.}
    \label{tab:primitive}
    \vspace*{-.6\baselineskip}
\end{table}
\noindent \textbf{Ablation Study:} The impact of the components proposed in CAD-SIGNet is assessed in Table~\ref{tab:ablation} in terms of F1 scores and CAD reconstruction metrics (IR, mean, and median CD). In the first row of this Table, the hybrid sampling described in Section~\ref{sec:train-inference} is ablated, thus selecting tokens with \textit{top-1} probabilities. It can be observed that this results in a drop of performance in terms of CD distances and IR while maintaining similar sketch and extrusion scores. A similar trend is observed for the second row, where the SGA module is omitted in the layer-wise cross-attention described in Section~\ref{sec:cross-attention}. This suggests that the hybrid sampling and the SGA module are especially important to obtain valid and accurate final CAD reconstructions. Finally, the third row reports the results when the layer-wise cross-attention defined in Eq.~(\ref{eq:cross-attention}) is not considered. In this case, each CAD language embedding $\mathbf F_b^c$ cross-attends to only the point cloud embedding $\mathbf F_B^v$ from the last block $B$. In other words, this  experiment omits passing the information from intermediate geometric embedding to the CAD language one. Despite generating only valid CAD reconstructions, we observe a drastic performance drop in all other metrics using this strategy compared to the proposed layer-wise cross-attention. 
Visual results of the ablation experiments are provided in supplementary materials. 


\begin{table}
\vspace*{-.5\baselineskip}
\resizebox{\linewidth}{!}{
\begin{tabular}{lccccc}
\hline
\multirow{2}{*}{Model} & \multicolumn{2}{c}{F1 Score}& \multicolumn{1}{c}{\multirow{2}{*}{\begin{tabular}[c]{@{}c@{}}Median  CD$\downarrow$\\$(\times 10^3)$\end{tabular}}} & \multicolumn{1}{c}{\multirow{2}{*}{\begin{tabular}[c]{@{}c@{}}Mean  CD$\downarrow$\\$(\times 10^3)$\end{tabular}}} & \multirow{2}{*}{IR$\downarrow$}\\
                       & Sketch$\uparrow$& Extrusion$\uparrow$& \multicolumn{1}{c}{}                                                                      & \multicolumn{1}{c}{}                                                                    &                     \\ \hline
Ours w/o Hybrid Samp.&                      75.30&                       \textbf{92.97}&                                                                                           0.291&                                                                                         6.784&                     5.02\\
 Ours w/o SGA & 75.13& 92.49& 0.289& 4.995&2.18\\
 Ours w/o Layer-wise CA & 45.89& 84.53& 76.40& 122.7&\textbf{0}\\ 
 \textbf{CAD-SIGNet (Ours)}& \textbf{75.94}& 92.72& \textbf{0.283}& \textbf{3.432}&                     0.88\\
 \hline \end{tabular}
}
\vspace*{-.5\baselineskip}
\caption{Ablation study. Sketch and extrusion F1 scores. Median, Mean CD, and IR metrics.}
    \label{tab:ablation}
    \vspace*{-.6\baselineskip}
\end{table}

\vspace*{-.4\baselineskip}
\subsection{Conditional Auto-Completion from User Input}
\label{subsec:ac}

CAD-SIGNet's auto-regressive nature enables it to complete the next design steps given a user input and a complete point cloud. To showcase this scenario, the same model trained for full design history recovery is used. During inference, the ground-truth tokens of the first extrusion and sketch are provided, and the task is to predict the next tokens of the CAD sequence given the complete point cloud.


\vspace{0.05cm}
\noindent \textbf{Baseline Methods:} To the best of our knowledge, there is no existing method capable of achieving the aforementioned task. DeepCAD~\cite{Wu_2021_ICCV} and MultiCAD~\cite{multicad} are not suitable candidates for this task due to their feed-forward nature. For the sake of comparison, we adapt two auto-regressive generative models, namely SkexGen~\cite{xu2022skexgen}, and HNC~\cite{hnc}.
Similarly to DeepCAD~\cite{Wu_2021_ICCV}, we train a PointNet++~\cite{pnet++} to encode the point cloud into the latent space learned by SkexGen~\cite{xu2022skexgen}, and HNC~\cite{hnc} on CAD language. Note that these adapted baselines were not subject to optimization. More details on how these methods are adapted are provided in the supplementary materials.

\vspace{0.05cm}
\noindent \textbf{Metrics:} The auto-completion performance is evaluated in terms of final CAD reconstructions. This is measured by the IR introduced in Section~\ref{sec:design-history} and another measure based on the CD. The latter is given by computing: (1) $\text{CD}_{\text{pred}}^{\text{gt}}$ which is the CD between the CAD reconstruction of the completed sequence and the ground-truth CAD model, (2) $\text{CD}_{\text{in}}^{\text{gt}}$ which is the CD between the CAD reconstruction of the user input sequence and the ground-truth, (3) the ratio of the two distances $\text{CD}_{\text{pred}}^{\text{gt}} / \text{CD}_{\text{in}}^{\text{gt}}$. This ratio allows for assessing whether the completed sequence resulted in a better final CAD reconstruction with respect to the user input. In order to reflect the distribution of this measure, the three quartiles Q1, Q2 (i.e., median), and Q3 are reported.  

\vspace{0.05cm}
\noindent \textbf{Results:} Table~\ref{tab:autocom} compares the results of CAD-SIGNet when using hybrid sampling and without, to the adapted baselines based on HNC~\cite{hnc} and SkexGen~\cite{xu2022skexgen}
. It can be observed that all the quartile values of the CD ratio for SkexGen baseline are very close to $1$ which indicates that the completed sequences resulted in a final CAD reconstruction that is close to the one of the user input. 
Notably, CAD-SIGNet achieved low Q1 and Q2 values of $0.054$ and $0.325$, respectively, showing that it improved the user input by a large margin on half of the testing samples. These observations are consistent with the visual results reported in Figure~\ref{fig:Autocompletion}. Moreover, discarding the hybrid sampling yields in a lower performance in all metrics but still outperforms both HNC and SkexGen baselines. Finally, the SGA module provides a noticeable improvement on the median CD ratio which further highlights its importance. 


\begin{table}[H]
\vspace*{-.5\baselineskip}
\resizebox{\linewidth}{!}{\begin{tabular}{lcccc}
\hline
\multirow{2}{*}{Model}& \multicolumn{3}{c}{CD Ratio} & \multirow{2}{*}{IR$\downarrow$}\\
                                & Q1$\downarrow$& Q2 (Median)$\downarrow$& Q3$\downarrow$&                                          \\ \hline
SkexGen-Baseline~\cite{xu2022skexgen}                         &               0.987&                    1.000&               1.035& 2.04\\
HNC-Baseline~\cite{hnc}                                &               0.437&                    1.015&               2.589&                                          8.85\\
Ours w/o Hybrid Samp.  & 0.096& 0.696& 1.096&4.40\\
 Ours w/o SGA& 0.060& 0.458& \textbf{0.992}&0.91\\
\textbf{CAD-SIGNet (Ours)}                           & \textbf{0.054}& \textbf{0.325}& 0.995& \textbf{0.65}\\ \hline
\end{tabular}}
\vspace*{-.5\baselineskip}
\caption{Conditional auto-completion from user input and point clouds on DeepCAD~\cite{Wu_2021_ICCV} dataset. CD ratio Quartiles and IR results.}
    \label{tab:autocom}
    \vspace*{-.2\baselineskip}
\end{table}

\begin{figure}
\centering
    \includegraphics[width=.92\linewidth]{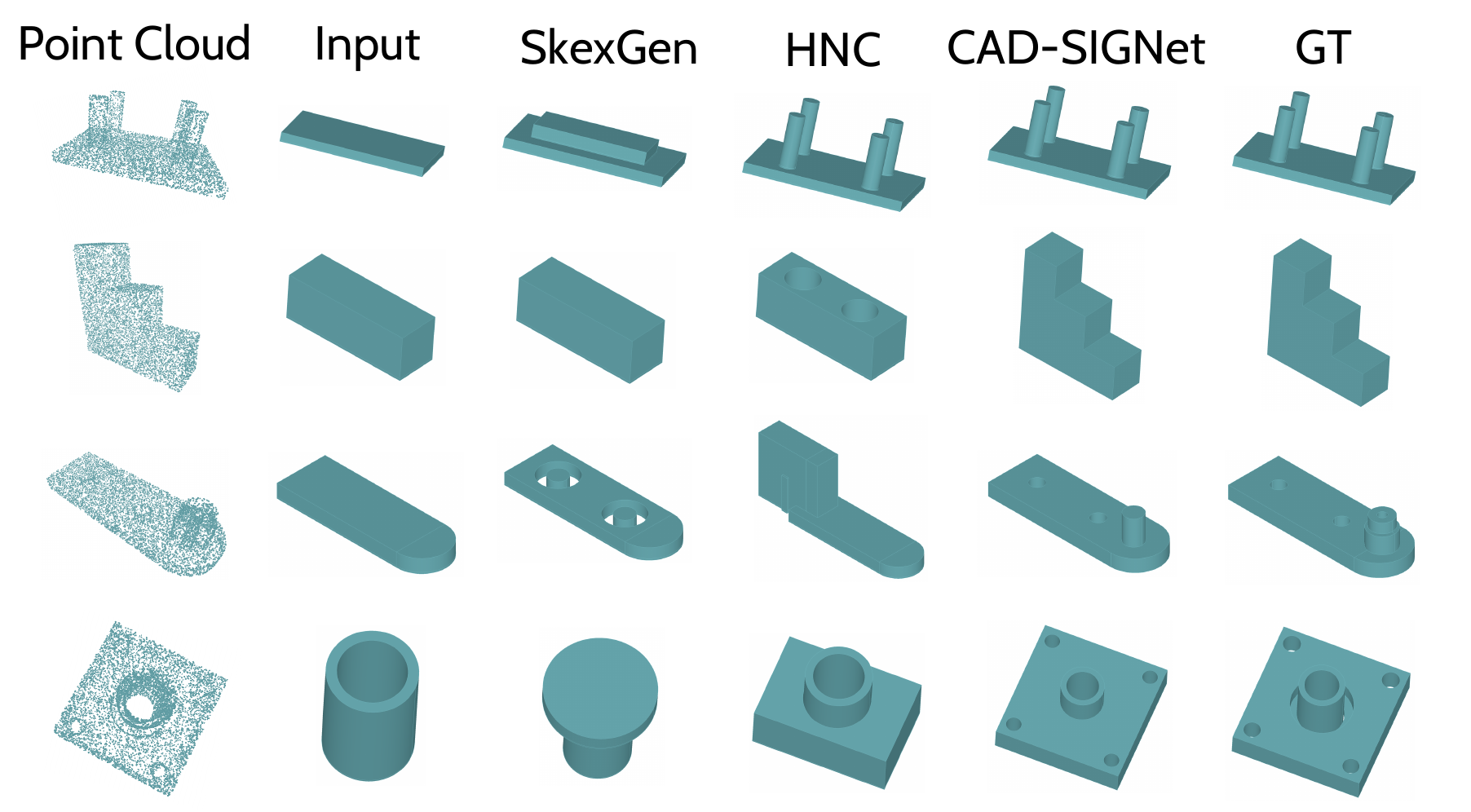}
        \setlength{\belowcaptionskip}{-10pt}
        \vspace*{-.5\baselineskip}
    \caption{Visual results for auto-completion from user input on DeepCAD~\cite{Wu_2021_ICCV} dataset. From left to right, input point cloud, CAD model reconstruction from user input CAD sequence, SkexGen~\cite{xu2022skexgen}, HNC~\cite{hnc}, CAD-SIGNet (ours), and ground-truth.}
    \label{fig:Autocompletion}
    \vspace*{-.2\baselineskip}
\end{figure}     
\vspace*{-.4\baselineskip}



\vspace*{-.9\baselineskip}
\subsection{Applications of CAD-SIGNet}\label{sec:application_cad_signet}

In this section, we highilight the applicability of CAD-SIGNet in a real-world scenario of reverse engineering. 
\vspace{0.04cm}
\noindent \textbf{Cross-Dataset Experiment on Fusion360}:
Following the protocol outlined in MultiCAD~\cite{multicad}, a cross-dataset experiment is performed on the Fusion360 dataset~\cite{willis2021fusion}. Results presented in Table~\ref{tab:fusion_icd} shows that CAD-SIGNet outperforms both MultiCAD~\cite{multicad} and DeepCAD~\cite{Wu_2021_ICCV} by a significant margin. Figure~\ref{fig:vis-res} (right) shows some visual comparison of the reconstructed CAD models from the predicted CAD sequences of CAD-SIGNet and DeepCAD~\cite{Wu_2021_ICCV}. The results indicate that CAD-SIGNet achieves better 3D reconstruction quality in comparison to DeepCAD~\cite{Wu_2021_ICCV}.
\begin{table}[ht]
\centering
\vspace*{-.5\baselineskip}
\resizebox{0.9\linewidth}{!}{
\begin{tabular}{lcc}
\hline
Model                      & \multicolumn{1}{c}{IR$\downarrow$} & \multicolumn{1}{c}{Median CD ($\times10^3)$$\downarrow$} \\ \hline
DeepCAD~\cite{Wu_2021_ICCV}                   &                        25.17&                               89.2\\
MultiCAD~\cite{multicad}                &                        16.52&                               42.2\\
\textbf{CAD-SIGNet (Ours)} &                        \textbf{1.83}&                               \textbf{1.15}\\ \hline
\end{tabular}
}
\vspace*{-.5\baselineskip}
\caption{Cross-dataset experiment on design history recovery from point clouds on Fusion360~\cite{willis2021fusion} dataset. The results for~\cite{multicad,Wu_2021_ICCV} are the ones reported in~\cite{multicad}.}
\label{tab:fusion_icd}
\vspace*{-.6\baselineskip}
\end{table}

\vspace{0.04cm}
\noindent \textbf{Design History Recovery from Realistic 3D Scans}:
The reported results on DeepCAD dataset~\cite{Wu_2021_ICCV} are obtained by applying the model on point clouds sampled on CAD meshes. However, in a real-world scenario of reverse engineering, we aim to reverse engineer 3D scans which are prone to 3D scanning artifacts. The CAD-SIGNet model trained on DeepCAD dataset is tested on this setting by performing a cross-dataset testing
. The CC3D dataset consists of 50k+ realistic 3D scans with their corresponding CAD models. Figure~\ref{fig:vis-res} shows some qualitative results of CAD-SIGNet compared to DeepCAD. Despite not being trained on such scan data, CAD-SIGNet succeeds in reconstructing promising CAD reconstructions. On the test set of CC3D dataset, composed of $5570$ samples, we report a median CD of $2.398$ and an IR of $2.39$ compared to a median CD of $263.56$ and an IR of $12.73$ achieved by DeepCAD. 

\noindent \textbf{User Controlled Reverse Engineering}: In a real-world reverse engineering scenario, it is not only desirable to generate the correct CAD sequence from a given point cloud, but to provide the user with a choice over every design step~\cite{hnc}. Towards this direction, one can further extend the hybrid sampling strategy to generate either multiple sketch planes for the extrusion steps or multiple sketches or loops from one single sketch plane. Since sketch sequence generation relies on the points laying close to the predicted sketch planes, changing sketch planes can result in a new sketch. The right panel of Figure~\ref{fig:teaser} shows different recommended design paths generated by our method a user can interactively follow along the design process.

\vspace*{-.4\baselineskip}
\section{Conclusion}
\label{sec:conclusion}
\vspace{-.2\baselineskip}

In this paper, we propose, CAD-SIGNet, an auto-regressive architecture designed for recovering the design history of a CAD model given a point cloud. This history is represented as a sequence of sketch-and-extrusion sequences. Leveraging its auto-regressive nature, CAD-SIGNet reconstructs a CAD design history from the input point cloud, simultaneously offering a range of plausible design alternatives. Through multi-modal transformer blocks of layer-wise cross-attention, the information is passed between CAD language and point cloud embedding. Notably, the incorporation of the SGA module enhances the reconstruction of fine-grained details in the sketches. As future works, we believe that selecting subsets of points using SGA could help overcome the high computational costs associated with large point clouds, currently limited to $8192$ points. While our work only considers the extrusion operation, CAD-SIGNet could be adapted for other operations.

\vspace{-.4\baselineskip}
\section{Acknowledgements}
\vspace{-.2\baselineskip}
The present project is supported by the National Research Fund, Luxembourg under the BRIDGES2021/IS/16849599/FREE-3D and
IF/17052459/CASCADES.

{
    \small
    \bibliographystyle{ieeenat_fullname}
    \bibliography{main}
}
\setcounter{page}{1}
\maketitlesupplementary

In this document, we provide more details on the used CAD sequence representation (Section~(\ref{sec:CADseq})), the experimental setup (Section~(\ref{sec:exp})), and the evaluation (Section~(\ref{sec:eval})).

\vspace{-.4\baselineskip}
\section{CAD Sequence Representation Details}
\label{sec:CADseq}
In this section, further details on the CAD sequence representation are provided. Table~\ref{tab:token} provides an overview of the tokens and their value ranges within the CAD sequence representation $\BC$.  Each extrusion sequence is composed of $11$ tokens specified in the following order: $\{d^+,d^-,\tau_x,\tau_y,\tau_z,\theta,\phi,\gamma,\sigma,\beta,e_e\}$. On the other hand, a sketch sequence can be defined by a variable number of tokens and follows the hierarchical structure mentioned in~\cite{xu2022skexgen}. As described in Section~\color{red}3 \color{black} of the main paper, a sketch sequence consists of curves represented by a sequence of 2D point coordinates $(p_x, p_y)$. Each curve type is formulated using the following parameters:
 \begin{itemize}
     \item Line: Start and End point.
     \item Arc: Start, Mid, and End point.
     \item Circle: Center and top-most point.
 \end{itemize}
 Following~\cite{Wu_2021_ICCV}, apart from the non-numerical tokens$~\{e_c,e_l,e_f,e_s,\beta,e_e,cls,end,pad\}$, all the other tokens in the CAD sequence $\BC$ are quantized to $8$ bits. Notably, the first $11$ classes are reserved for the non-numerical tokens resulting in a total of $d_t=266~(2^8+11)$ classes. Post quantization, each one dimensional token is augmented into two dimensions with a $\textit{pad}$ token. An example of a CAD sequence representation is depicted in Figure~\ref{fig:sec_rep}.

\begin{table}[]
\resizebox{\linewidth}{!}{\begin{tabular}{ccccc}
\hline
\begin{tabular}[c]{@{}c@{}}Sequence\\ Type\end{tabular}                        & \begin{tabular}[c]{@{}c@{}}Token\\ Type\end{tabular} & \begin{tabular}[c]{@{}c@{}}Token\\ Flags\end{tabular} & \begin{tabular}[c]{@{}c@{}}Token\\ Value\end{tabular} & Description                                                                                \\ \hline
\multicolumn{1}{l}{}                                                           & $pad$                                                & 11                                                    & $0$                                                   & Padding Token                                                                              \\
\multicolumn{1}{l}{}                                                           & $cls$                                                & 0                                                     & $1$                                                   & Start Token                                                                                \\
\multicolumn{1}{l}{}                                                           & $end$                                                & 0                                                     & $1$                                                   & End Token                                                                                  \\ \hline
\multicolumn{1}{l}{\multirow{5}{*}{}}                                          & $e_s$                                                & 0                                                     & $2$                                                   & End Sketch                                                                                 \\
\multicolumn{1}{l}{}                                                           & $e_f$                                                & 0                                                     & $3$                                                   & End Face                                                                                   \\
\multicolumn{1}{l}{}                                                           & $e_l$                                                & 0                                                     & $4$                                                   & End Loop                                                                                   \\
\multicolumn{1}{l}{}                                                           & $e_c$                                                & 0                                                     & $5$                                                   & End Curve                                                                                  \\
\multicolumn{1}{l}{}                                                           & $(p_x,p_y)$                                          & 0                                                     & $\Iintv{11,266}^2$                                    & Coordinates                                                                                \\ \hline
\multirow{11}{*}{\begin{tabular}[c]{@{}c@{}}Extrusion\\ Sequence\end{tabular}} & $d^+$                                                & 1                                                     & $\Iintv{11,266}$                                      & \begin{tabular}[c]{@{}c@{}}Extrusion Distance Towards \\ Sketch Plane Normal\end{tabular}  \\ \cline{2-4}
                                                                               & $d^-$                                                & 1                                                     & $\Iintv{11,266}$                                      & \begin{tabular}[c]{@{}c@{}}Extrusion Distance Opposite \\ Sketch Plane Normal\end{tabular} \\ \cline{2-4}
                                                                               & $\tau_x$                                             & 2                                                     & $\Iintv{11,266}$                                      & \multirow{3}{*}{Sketch Plane Origin}                                                       \\
                                                                               & $\tau_y$                                             & 3                                                     & $\Iintv{11,266}$                                      &                                                                                            \\
                                                                               & $\tau_z$                                             & 4                                                     & $\Iintv{11,266}$                                      &                                                                                            \\ \cline{2-4}
                                                                               & $\theta$                                             & 5                                                     & $\Iintv{11,266}$                                      & \multirow{3}{*}{Sketch Plane Orientation}                                                  \\
                                                                               & $\phi$                                               & 6                                                     & $\Iintv{11,266}$                                      &                                                                                            \\
                                                                               & $\gamma$                                             & 7                                                     & $\Iintv{11,266}$                                      &                                                                                            \\ \cline{2-4}
                                                                               & $\sigma$                                             & 8                                                     & $\Iintv{11,266}$                                      & Sketch Scaling Factor                                                                      \\
                                                                               & $\beta$                                              & 9                                                     & $\{7,8,9,10\}$                                         & Boolean (New, Cut, Join, Intersect)                                                        \\
                                                                               & $e_e$                                                & 10                                                    & $6$                                                   & End Extrude                                                                                \\ \hline
\end{tabular}}
\caption{Description of different tokens used in our CAD language representation.}
\label{tab:token}
\vspace{-.8\baselineskip}
\end{table}

\begin{figure*}
\centering
    \includegraphics[width=1\linewidth]{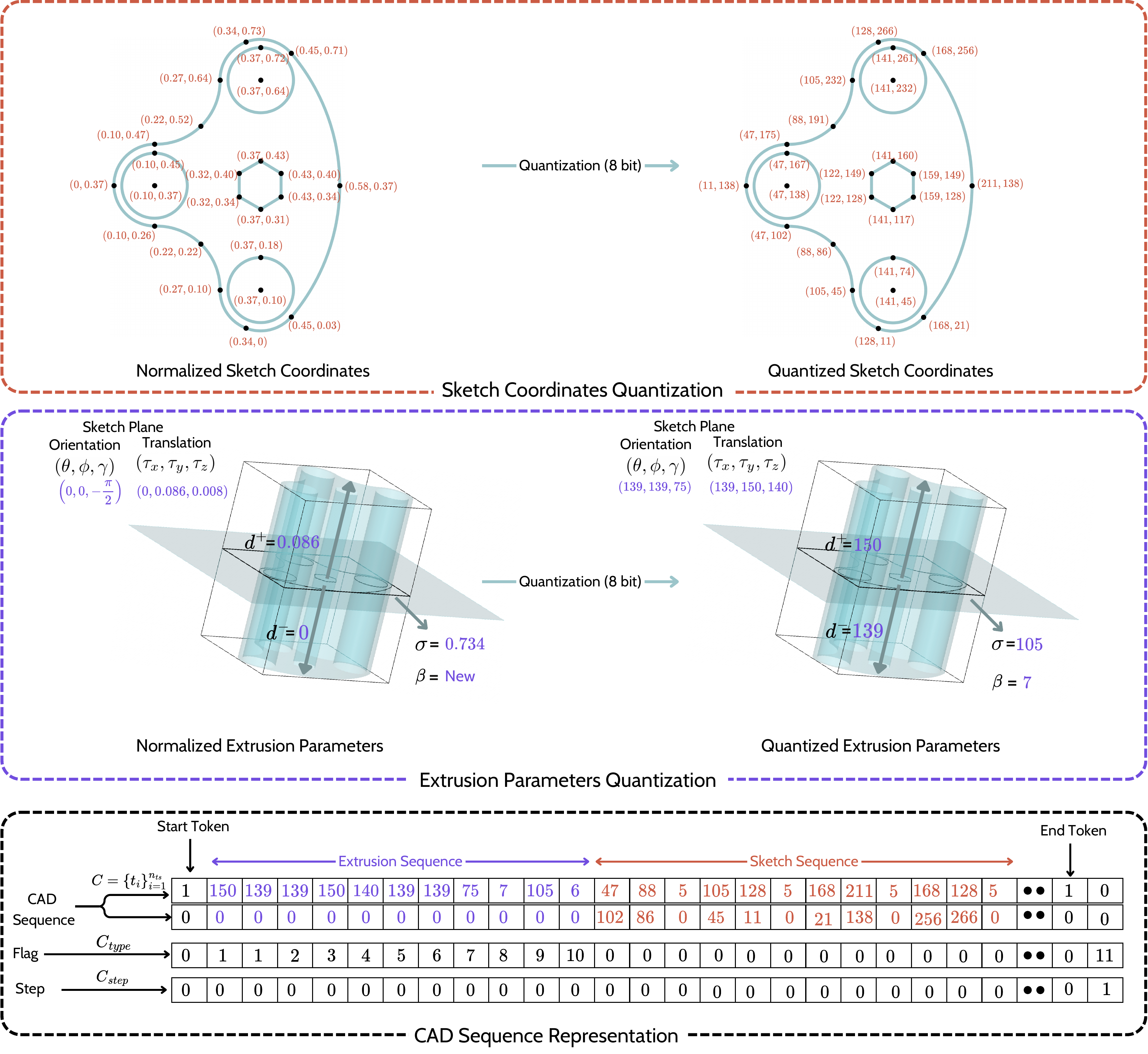}
        \setlength{\belowcaptionskip}{-10pt}
        \vspace*{-.9\baselineskip}
    \caption{Example of a CAD Sequence Representation. The top and middle panels show the $8$-bit quantization process of sketch and extrusion parameters respectively. The bottom panel depicts the construction of the sequence from the different tokens.}
    \label{fig:sec_rep}
    \vspace{-.8\baselineskip}
\end{figure*}



\section{Additional Details on Experiments}
\label{sec:exp}

In this section, further details on the experimental procedure are provided.
\subsection{Data Preprocessing Details}

During preprocessing, each sketch element (faces, loops, and curves) is reordered from the original sequence order. We follow the approach of~\cite{Wu_2021_ICCV}, in which sketch elements are reordered according to their bounding box bottom-left position in ascending order. Furthermore, curves are oriented in a counter-clockwise direction as in~\cite{Wu_2021_ICCV}. Similar to~\cite{Wu_2021_ICCV}, at most $10$ extrusions, are considered for our experiments, resulting in a maximum CAD sequence length of $n_{ts}=273$. Given the variable number of extrusions within a CAD sequence $\BC$, padding tokens (\textit{pad, pad}) are appended at the end of the sequences during training for batch processing. This ensures that every sequence in a batch has a length of $273$.

\subsection{Training Details}
The AdamW~\cite{adamw} optimizer is used during training with a learning rate of $0.001$. Additionally, an ExponentialLR scheduler is applied to adjust the learning rate during training, with a decay factor of $\gamma=0.999$. The dropout rate is set to $0.1$. For the LFA~\cite{randlanet} k-NN feature aggregation, the number of neighbors is set to $4$. In the first two multi-modal transformer blocks, cross-attention is not used between the CAD sequence and point embedding. This design choice is made to prioritize the learning of the intra-modality relationship in early layers. The training time is approximately $6$ days for $150$ epochs.

\subsection{More on Design History Evaluation Metrics}
In the main paper, \textit{F1} scores on the extrusions and curve types are reported as a measure of the quality of the predicted CAD sequences. To compute the \textit{F1} scores, the positions of the End Sketch ($e_s$) and End Extrude ($e_e$) tokens are initially identified for each ground truth and predicted CAD sequence. This allows us to divide each sequence into a list of sketches and extrusions. The extrusion \textit{F1} score is computed on the numbers of ground truth and predicted extrusion sequences. 
To compute the \textit{F1} scores for each curve type, the procedure described in Algorithm~\ref{alg:calc_metrics} is used. In this algorithm, the loops of the ground truth and predicted sketches of the same step are matched using the \textit{match\_entity\_list} function described in Algorithm~\ref{alg:match_loop_curve}. The \textit{match\_entity\_list} function employs a Hungarian matching~\cite{hungarian-matching} to establish the correspondences between two lists of loops. The cost associated with matching two loops is defined as the sum of the Euclidean distances between their respective bounding box bottom-left and top-right corners. A similar matching strategy is extended for the curves within the matched loops. Finally, the list of matched curve pairs from all the sketches is used to compute the curve type \textit{F1} scores. 

\noindent Parameter accuracy introduced in~\cite{Wu_2021_ICCV} is omitted in our work.
This is because the CAD sequence representation in DeepCAD~\cite{Wu_2021_ICCV} differs significantly from ours, particularly in the curve parameterization. Attempting to transform predictions from one representation to another will propagate prediction errors, resulting in an unfair comparison. 
\begin{algorithm}
\SetAlgoNlRelativeSize{-1}
\KwData{$S_{g},S_{p}$} \tcp{\textcolor{commentColor}{\textnormal{List of ground-truth and predicted sketches.}}}
\KwResult{Recall, Primitve, F1 for curves.}
\texttt{n\_gt} $\leftarrow$ \textit{length}($S_{g}$)\\
\texttt{n\_pred} $\leftarrow$ \textit{length}($S_{p}$)\\
\texttt{n\_max} $\leftarrow$ \textit{max}(\texttt{n\_gt, n\_pred})\\
\tcp{\textcolor{commentColor}{\textnormal{Over or under-prediction of sketches.}}}
\If{\texttt{n\_gt} $\neq$ \texttt{n\_pred}}{
    Append \texttt{None} to $S_{g}$ or $S_{p}$ until their lengths become~\texttt{n\_max}.\\
}

\tcp{\textcolor{commentColor}{\textnormal{List of ground truth and predicted curve types.}}}
\texttt{y\_true} $\leftarrow$ []\\
\texttt{y\_pred} $\leftarrow$ []

\For{$i \leftarrow 1$ \KwTo $n\_max$}{
 \tcp{\textcolor{commentColor}{\textnormal{Match loops in the sketch.}}}
   \texttt{loop\_pair} $\leftarrow$ \textit{match\_entity\_list}(
   $S_g[i]$.\texttt{loopList},
   $S_p[i]$.\texttt{loopList}
   )\\
    \tcp{\textcolor{commentColor}{\textnormal{Match curves in the matched loop pairs.}}}
   \For{($l_g,l_p$) in \texttt{loop\_pair}}{
   \tcp{\textcolor{commentColor}{\textnormal{Get pairs of ground truth and predicted curves from matched loops.}}}
   $(c_g,c_p)$ $\leftarrow$ \textit{match\_entity\_list}(
   $l_g$.\texttt{curveList},
   $l_p$.\texttt{curveList}
   )\\

    append \textit{curve\_type}($c_g$) in \texttt{y\_true}\\
    append \textit{curve\_type}($c_p$) in \texttt{y\_pred}
   }
 
}
\texttt{recall} $\leftarrow$ \textit{multiclass\_recall}\texttt{(y\_true,y\_pred)}\\
\texttt{precision} $\leftarrow$ \textit{multiclass\_precision}\texttt{(y\_true,y\_pred)}\\
\texttt{f1} $\leftarrow$ \textit{multiclass\_f1}\texttt{(y\_true,y\_pred)}\\
\textit{return} \texttt{recall, precision, f1}
\caption{\textit{calculate\_metrics}}
\label{alg:calc_metrics}
\end{algorithm}

\noindent For evaluating the CAD reconstruction (obtained by Opencascade~\cite{opencascade}), the Chamfer Distance (CD)~\cite{chamfer_distance} is computed. 
This is achieved by uniformly sampling $8192$ points from the predicted and ground truth reconstructed CAD models. To ensure scale-invariance in CD computation, the models are normalized within a unit bounding box. Note that the CD can only be computed if the predicted sequence leads to a valid CAD model.
\vspace*{-.8\baselineskip}
\section{Additional Evaluation Details}
\label{sec:eval}

\begin{algorithm}
\SetAlgoNlRelativeSize{-1}
\KwData{$e_g,e_p$} \tcp{\textcolor{commentColor}{\textnormal{List of ground-truth and predicted entities of the same sketch index. Entity can be a loop or a curve.}}}
\KwResult{Matched Curve Pair}
\texttt{n\_gt} $\leftarrow$ \textit{length}($e_g$)\\
\texttt{n\_pred} $\leftarrow$ \textit{length}($e_p$)\\
\texttt{n\_max} $\leftarrow$ \textit{max}(\texttt{n\_gt, n\_pred})\\
\tcp{\textcolor{commentColor}{\textnormal{Over or under-prediction of entities.}}}
 \If{\texttt{n\_gt} $\neq$ \texttt{n\_pred}}{
    Append \texttt{None} to $e_{g}$ or $e_{p}$ until their lengths become~\texttt{n\_max}.\\
}

\tcp{\textcolor{commentColor}{\textnormal{Hungarian matching cost matrix.}}}
\texttt{cost} $\leftarrow []$\\

\For{$i \leftarrow 1$ \KwTo $n\_gt$}{
\For{$j \leftarrow 1$ \KwTo $n\_pred$}{
\If{$e_g[i]$ and $e_p[i]$ is not \texttt{None}}{
   \texttt{cost}$[i][j]$=\textit{bbox\_distance}($e_g[i],e_p[j]$)
}
\Else{
   \texttt{cost}$[i][j]$=$\infty$
}
}
}
\texttt{matched\_entity\_pair}=\textit{hungarian\_matching(\texttt{cost})}\\
\textit{return} \texttt{matched\_entity\_pair}
\caption{\textit{match\_entity\_list}}
\label{alg:match_loop_curve}
\end{algorithm}
\vspace{-.6\baselineskip}
In this section, more results from the different experiments are shown.
\vspace{-.8\baselineskip}
\subsection{Model Parameters Comparison}
\vspace{-.6\baselineskip}
Table~\ref{tab:param_count} shows the number of parameters required for each of the networks presented in the main paper. CAD-SIGNet has the lowest number of parameters compared to other baselines.

\begin{table}[H]
\centering
\begin{tabular}{lc}
\hline
Model                & \#Parameters \\ \hline
DeepCAD~\cite{Wu_2021_ICCV}+PointNet++~\cite{pnet++}   & 7.4M       \\
SkexGen~\cite{xu2022skexgen} + PointNet++~\cite{pnet++} & 18.7M      \\
HNC~\cite{hnc} + PointNet++~\cite{pnet++}     & 58.4M      \\
\textbf{CAD-SIGNet (Ours)}           & \textbf{6.1M}       \\ \hline
\end{tabular}
\caption{Total number of parameters of CAD-SIGNet compared to different baseline models.}
\label{tab:param_count}
\end{table}


\vspace*{-\baselineskip}
\subsection{More Details on Design History Recovery}

\begin{table*}[ht]
\resizebox{\linewidth}{!}{
\begin{tabular}{lccc|ccc|ccc|lll}
\hline
\multirow{2}{*}{Model} & \multicolumn{3}{c|}{Line}                                                                                                                                                              & \multicolumn{3}{c|}{Arc}                                                                                                                                                               & \multicolumn{3}{c|}{Circle}                                                                                                                                                            & \multicolumn{3}{c}{Extrusion}                                                                                                                                                                                                                      \\
                       & Recall& Precision& F1& Recall& Precision& F1& Recall& Precision& F1& \multicolumn{1}{c}{Recall} & \multicolumn{1}{c}{Precision} & \multicolumn{1}{c}{F1} \\ \hline
DeepCAD~\cite{Wu_2021_ICCV} + PointNet++~\cite{pnet++} & 69.86                                                       & 72.40                                                          & 68.37                                                   & 12.53                                                       & 15.21                                                          & 12.89                                                   & 59.61                                                       & 61.95                                                          & 58.82                                                   & 81.74                                                                           & 94.87&                                                                              86.88\\
 DeepCAD~\cite{Wu_2021_ICCV} + LFA~\cite{randlanet}& 66.26& 71.09& 65.04& 4.07& 6.19& 4.41& 47.45& 50.49& 46.76& 79.63& 92.48&82.90\\ 
 Ours w/o Hybrid Samp.& 76.49& 80.12& 75.36& 26.90& 31.50& 27.45& 71.53& 71.78& 69.83& 93.98& \textbf{95.55}&\textbf{92.97}\\
 Ours w/o SGA & 77.07& \textbf{82.13}& 76.93& 26.12& 31.39& 26.89& 67.1& 69.08& 66.58& 94.17& 94.7&92.5\\
 Ours w/o Layer-wise CA& 56.63& 71.07& 56.99& 0.73& 1.37& 0.800& 18.34& 26.07& 19.97& 84.81& 92.96&84.53\\
 \textbf{CAD-SIGNet (Ours)}& \textbf{77.76}& 81.35& \textbf{77.31}& \textbf{27.67}& \textbf{33.01}& \textbf{28.65}& \textbf{72.07}& \textbf{72.22}& \textbf{70.36}& \textbf{94.26}& 94.93 & 92.72\\
  \hline
\end{tabular}
}
\caption{Recall, Precision, and F1 scores for the baseline and ablated CAD-SIGNet for lines, arcs, circles, and extrusions. The results are on DeepCAD~\cite{Wu_2021_ICCV} dataset.}
    \label{tab:full_eval}
    \vspace{-.8\baselineskip}
\end{table*}

Table~\ref{tab:full_eval} shows supplementary results for the baseline DeepCAD~\cite{Wu_2021_ICCV} and various versions of CAD-SIGNet as mentioned in the ablation study in Section~\color{red}5 \color{black} of the main paper. The precision and recall scores used to compute the \textit{F1} scores are also included. \\
In the first and second row of Table~\ref{tab:full_eval}, we provide the results of DeepCAD~\cite{Wu_2021_ICCV} with two distinct point cloud encoders: the first with PointNet++~\cite{pnet++}, and the second using LFA~\cite{randlanet} modules used in CAD-SIGNet. The transition to LFA~\cite{randlanet} modules from PointNet++~\cite{pnet++} leads to a noticeable decline in recall, precision, and F1 scores for DeepCAD~\cite{Wu_2021_ICCV}. 
In CAD-SIGNet without the Layer-wise CA, the point embedding from the last LFA layer are used for cross-attention in the multi-modal transformer blocks. This modification leads to a drastic drop in performance. In contrast, our retained model, which leverages Layer-wise CA, outperforms both the baseline DeepCAD~\cite{Wu_2021_ICCV} and CAD-SIGNet without Layer-wise CA. 
Additionally, emphasizing the importance of SGA and Hybrid Sampling, we observe improvements in recall, precision, and F1 scores, especially for arcs and circles.\\
\noindent Figure~\ref{fig:sga_vis} shows some examples of predicted sketch instances and the corresponding sketches compared to the ground truth ones. Sample $1$ from this figure shows that a correct sketch instance prediction has led to the correct sketch prediction and hence the final predicted CAD model matches the ground truth. In samples $2$ and $3$, the initial sketch instance is correctly predicted but the corresponding sketches do not match the ground truth ones. However, in these examples, the network can still predict a CAD reconstruction similar in shape to the ground truth. Sample $4$ shows that despite an incorrect sketch instance prediction, the final predicted CAD model has the same shape as the ground truth. 
Sample $5$ shows an example in which the correct sketch instance was identified but the network was unable to predict the corresponding sketch sequence. Finally, sample $6$ shows that the network sometimes fails to capture some details from the input point cloud. \\
Figure~\ref{fig:supp_des_hist} and~\ref{fig:supp_cc3d} showcase more visual results of the reconstructed CAD models from DeepCAD~\cite{Wu_2021_ICCV} and CAD-SIGNet on the DeepCAD~\cite{Wu_2021_ICCV} and CC3D~\cite{cc3d} datasets, respectively. The top panel of both figures shows that CAD-SIGNet achieves better performance than DeepCAD~\cite{Wu_2021_ICCV}, even for models containing higher curve counts. The bottom panel showcases invalid models of both DeepCAD~\cite{Wu_2021_ICCV} and CAD-SIGNet. We observe that the invalid outputs generated by CAD-SIGNet are due to syntax errors in the predictions such as a line or an arc with the same start and end point.\\
Figure~\ref{fig:supp_ablation} displays several qualitative results for the ablated versions of CAD-SIGNet. From those samples, it can be observed that the retained model consistently outperforms its ablated versions (\ie discarding layer-wise CA, SGA, and hybrid sampling). 


\subsection{Auto-Completion from User Input Details}
As outlined in Section~\color{red}5.2 \color{black} of the main paper, Skexgen~\cite{xu2022skexgen} and HNC~\cite{hnc} are used as baselines for the conditional auto-completion task. As both were not originally designed for conditional auto-completion from point clouds, their training strategy had to be adapted. For SkexGen~\cite{xu2022skexgen}, PointNet++~\cite{pnet++} is used to predict $10$ pretrained codebooks from the input point cloud. In the case of HNC~\cite{hnc}, the controllable CAD model generation network is retrained with modifications. In addition to the initial sketch and extrusion sequence, we incorporate the full point cloud as input. PointNet++~\cite{pnet++} is used to learn a latent embedding from the input point cloud. The resulting point cloud embedding was then appended to the CAD sequence embedding and passed to the code-tree generator to predict the codes. The generated codes along with the CAD sequence and point cloud embedding are fed to the sketch and extrusion decoder to generate the CAD sequence. Teacher forcing~\cite{teacher-forcing} strategy is used during training along with cross-entropy loss. The input CAD sequence is masked from the ground truth CAD sequence during the loss calculation. 

\noindent During conditional auto-completion inference of CAD-SIGNet, HNC~\cite{hnc}, SkexGen~\cite{xu2022skexgen}, the initial ground truth sketch and extrusion sequence along with the input point cloud are provided. The subsequent CAD sequence is autoregressively generated until the end token is predicted. \\
It is important to highlight that the CAD sequence representation in HNC~\cite{hnc} and SkexGen~\cite{xu2022skexgen} uses $6$-bit quantization while as our representation uses $8$-bit quantization. The different quantization schemes lead to different user inputs. Hence, the results reported in Table~\color{red}4 \color{black} of the main paper were considering the ratio of CD distances with respect to the user input, rather than CD.  
\vspace{-.2\baselineskip}
\subsection{Performance on Complex Models}
\vspace{-.2\baselineskip}
The complexity of models that CAD-SIGNet can handle depends on the training dataset. We consider two defining attributes for complexity: the number of extrusions and the total number of curves in the sketches within a CAD model. 
The DeepCAD dataset~\cite{Wu_2021_ICCV} which offers the required annotations for CAD-SIGNet includes models with at most 10 extrusions and 44 curves. Figure~\ref{fig:complexity_ext} shows the variation of the median CD \wrt the number of extrusions and the total number of curves per CAD model. Unlike DeepCAD~\cite{Wu_2021_ICCV}, the complexity of the input model has minor impact on the performance of CAD-SIGNet.

\begin{figure*}
    \centering
    \includegraphics[width=1\linewidth]{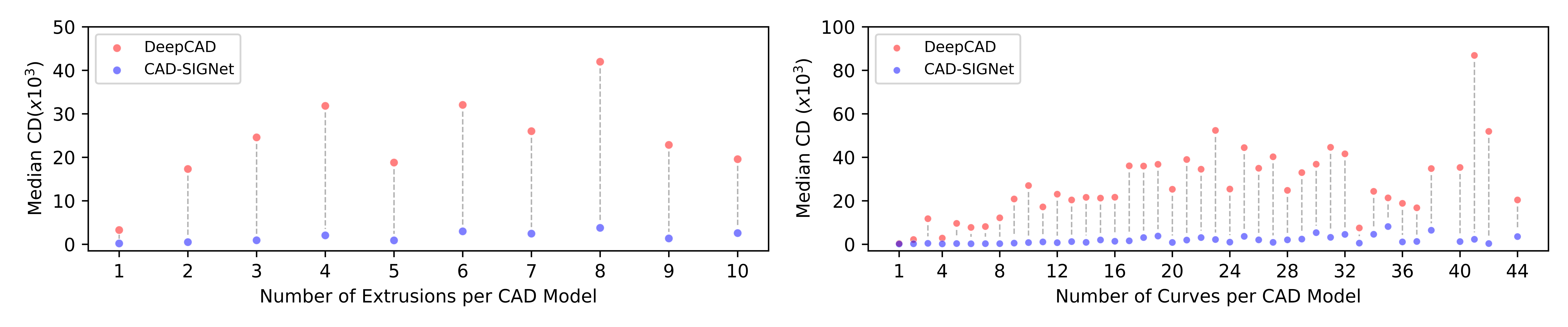}
    \caption{Comparison of median CD for reconstructed CAD models by CAD-SIGNet and DeepCAD~\cite{Wu_2021_ICCV} \wrt to the number of extrusions (\textbf{left}) and total number of curves (\textbf{right}) per CAD model.}
    \label{fig:complexity_ext}
\end{figure*}
\begin{figure*}
    \centering
    \includegraphics[width=1\linewidth]{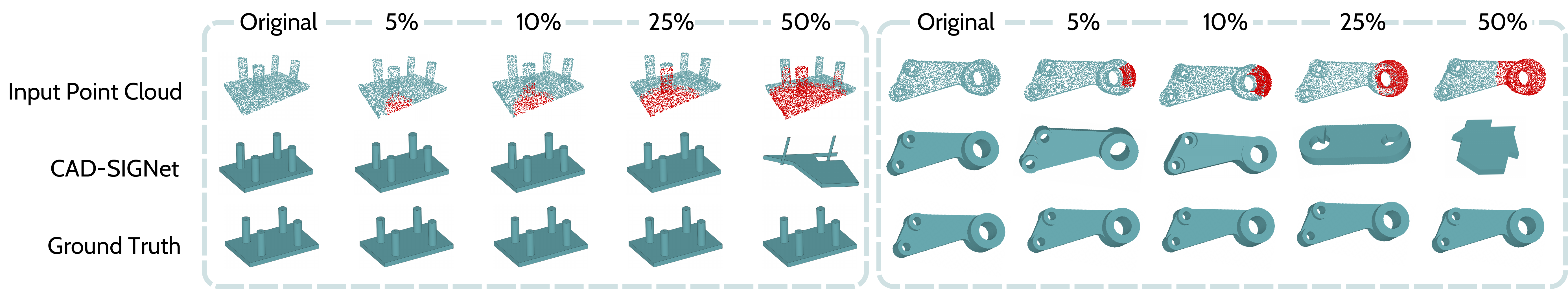}
    \caption{Performance of CAD-SIGNet under varying degrees of input point cloud occlusion. The percentage of missing points is indicated above each result. }
    \label{fig:pc_missing}
\end{figure*}
\vspace{-.2\baselineskip}
\subsection{Impact of Point cloud Quality}
The cross-dataset evaluation on the CC3D dataset~\cite{cc3d} in Section \ref{sec:application_cad_signet} aims at assessing the ability of CAD-SIGNet to reconstruct CAD models from point clouds that contain realistic scanning artifacts such as noise and self-occlusion.
 Furthermore, Figure~\ref{fig:pc_missing} shows some qualitative results from DeepCAD~\cite{Wu_2021_ICCV} dataset when a part of the input point cloud is missing. The top rows show the input point cloud with the missing points highlighted in red while the middle and bottom rows display the corresponding prediction from CAD-SIGNet and ground truth. We can observe that when a small portion of the point cloud is missing, CAD-SIGNet manages to recover a plausible solution. However, if a large portion of the input point cloud is missing, then CAD-SIGNet fails to capture the overall structure of the CAD model.

\vspace{-.2\baselineskip}
\subsection{Comparison with Point2Cyl}
In this section, we compare CAD-SIGNet with Point2Cyl~\cite{uy2022point2cyl}, another method addressing the 3D reverse engineering problem from point clouds. Figure~\ref{fig:point2cyl} shows some qualitative results between Point2Cyl~\cite{uy2022point2cyl} and CAD-SIGNet in terms of the output sketches and 3D reconstructions given an input point cloud. Notably, the shapes predicted by Point2Cyl~\cite{uy2022point2cyl} closely resemble the ground truth shapes. However, there are some major differences between our work and Point2Cyl~\cite{uy2022point2cyl}. 
Firstly, the sketches in Point2Cyl~\cite{uy2022point2cyl} are predicted as signed distance functions, and a marching square algorithm is applied to deduce the sketch. Such a strategy leads to a non-parametric form of the sketches. 
Secondly, the final model is a 3D mesh. This implies that the output model cannot be directly edited using CAD software, hence limiting the practical applications of this method. 
Finally, reconstructing the final model requires manual choice for the boolean operations between the different extrusion cylinders.

\begin{figure*}
    \centering
    \includegraphics[width=1\linewidth]{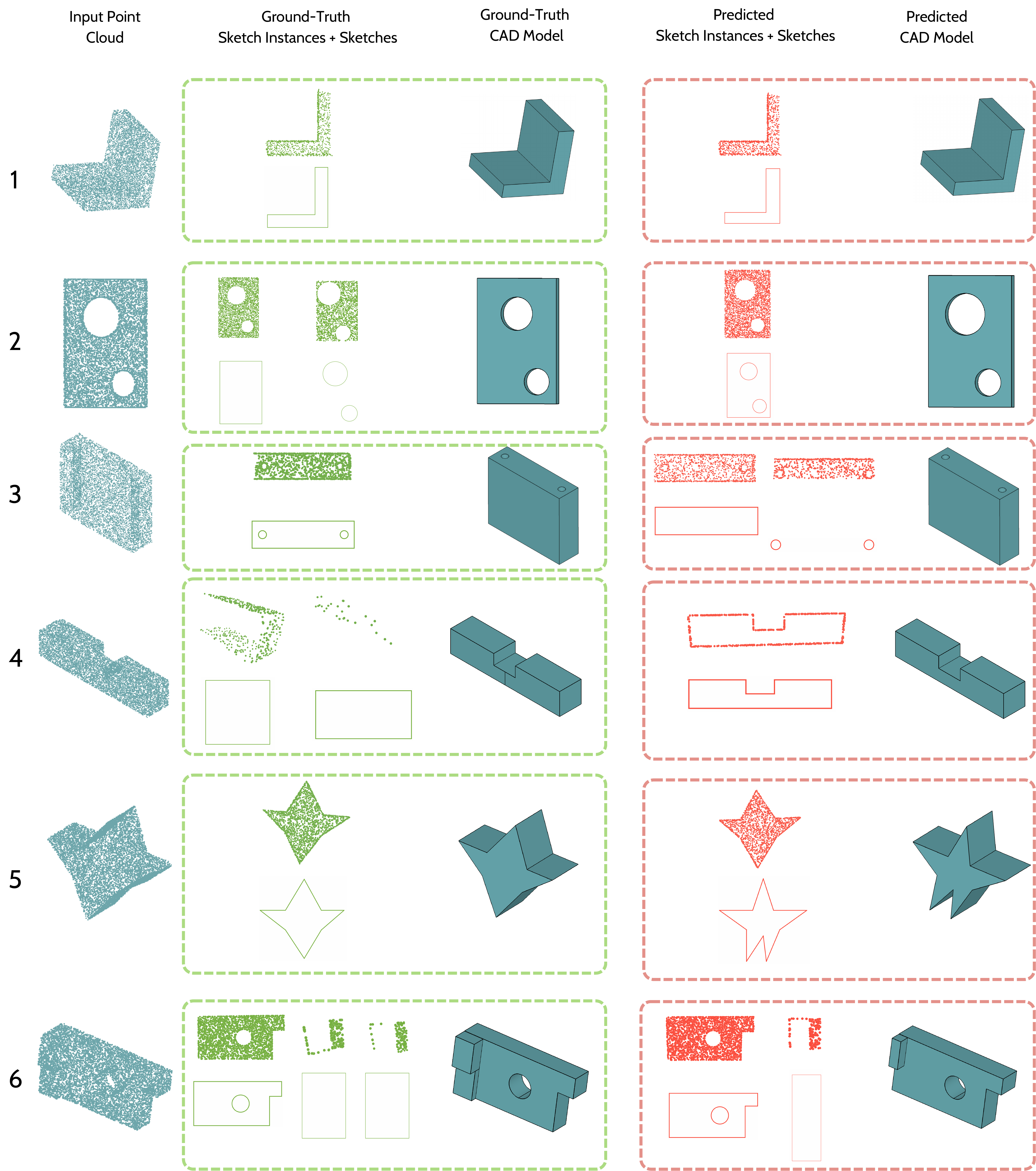}
    \caption{\textbf{Sketch Instances}. Ground truth (green) and predicted (red) sketch instances along with their corresponding sketches.  
    }
    \label{fig:sga_vis}
\end{figure*}


\begin{figure*}[t]
    \centering
    \includegraphics[height=1.25\linewidth]{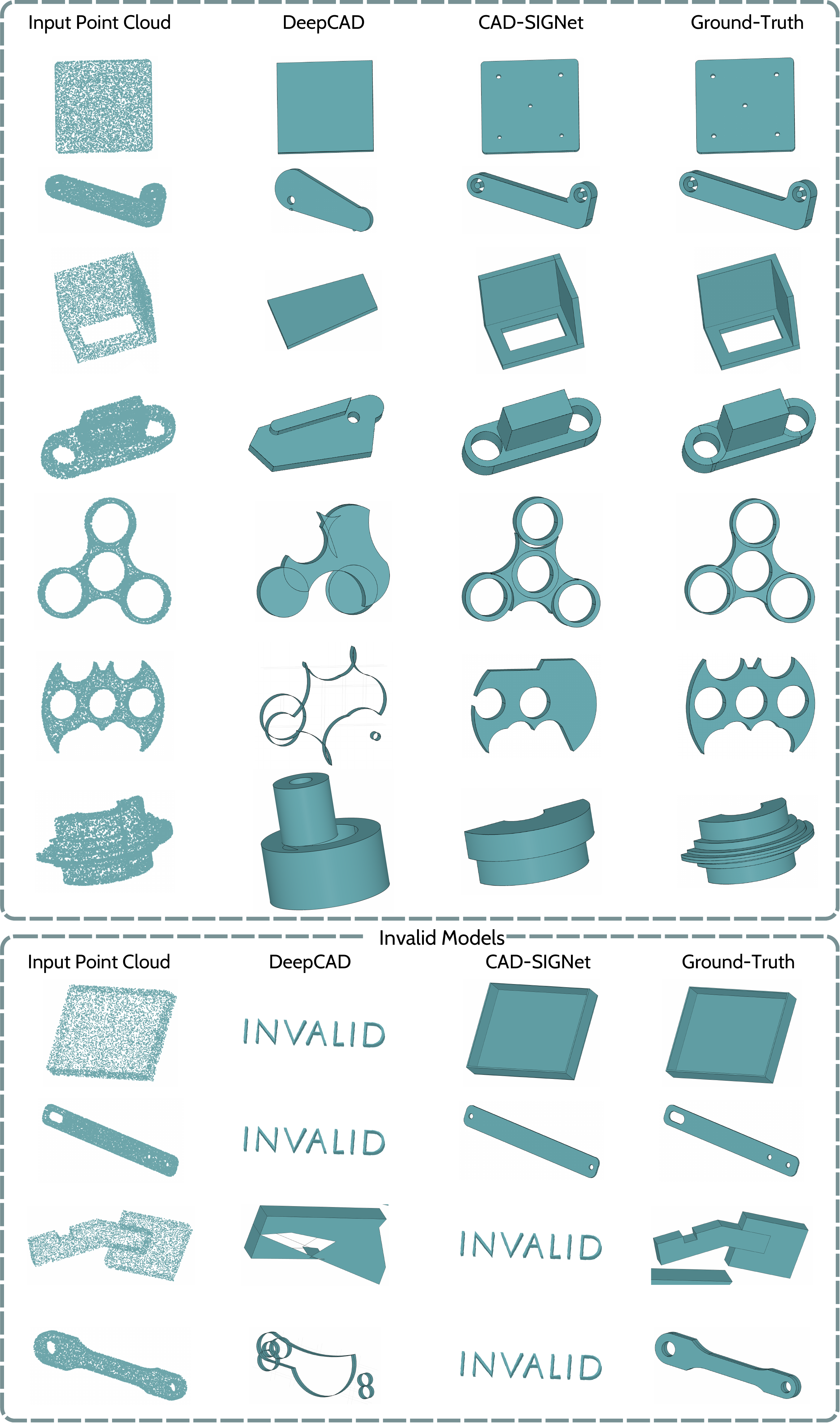}
    \caption{More qualitative results of the reconstructed CAD models from the input point clouds on the DeepCAD~\cite{Wu_2021_ICCV} dataset. The top panel shows examples of varying complexity. The bottom panel showcases invalid models, observed in both DeepCAD~\cite{Wu_2021_ICCV} and CAD-SIGNet.}
    \label{fig:supp_des_hist}
\end{figure*}

\begin{figure*}[t]
    \centering
    \includegraphics[height=1.25\linewidth]{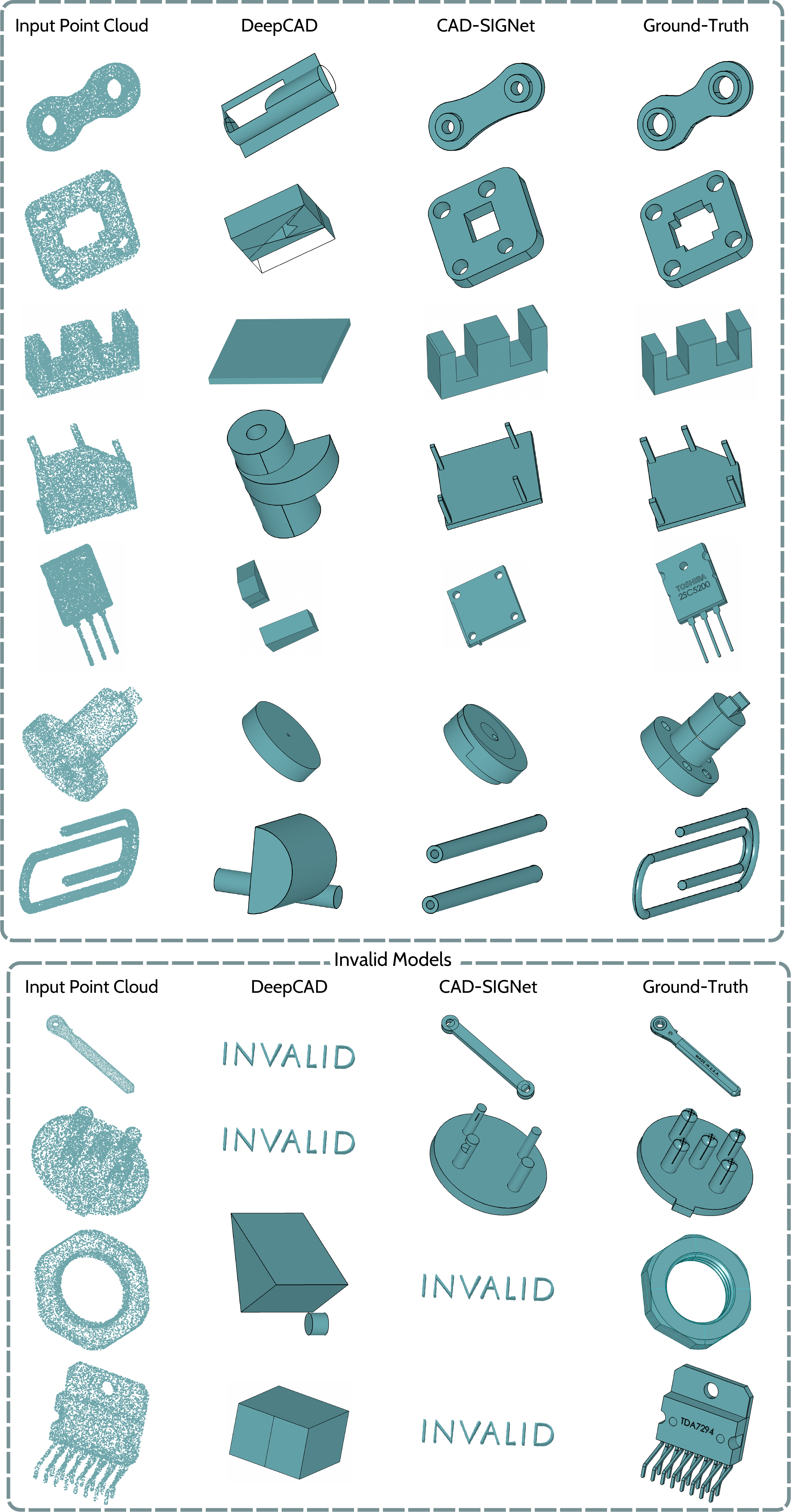}
    \caption{More qualitative results of the reconstructed CAD models from the CAD sequences predicted from input scans on CC3D~\cite{cc3d} dataset. Both CAD-SIGNet and DeepCAD~\cite{Wu_2021_ICCV} are trained on the DeepCAD~\cite{Wu_2021_ICCV} dataset. The top panel shows examples of varying complexity. The bottom panel showcases invalid models, observed in both DeepCAD~\cite{Wu_2021_ICCV} and CAD-SIGNet.}
    \label{fig:supp_cc3d}
\end{figure*}

\begin{figure*}[ht]
    \centering
    \includegraphics[width=\linewidth]{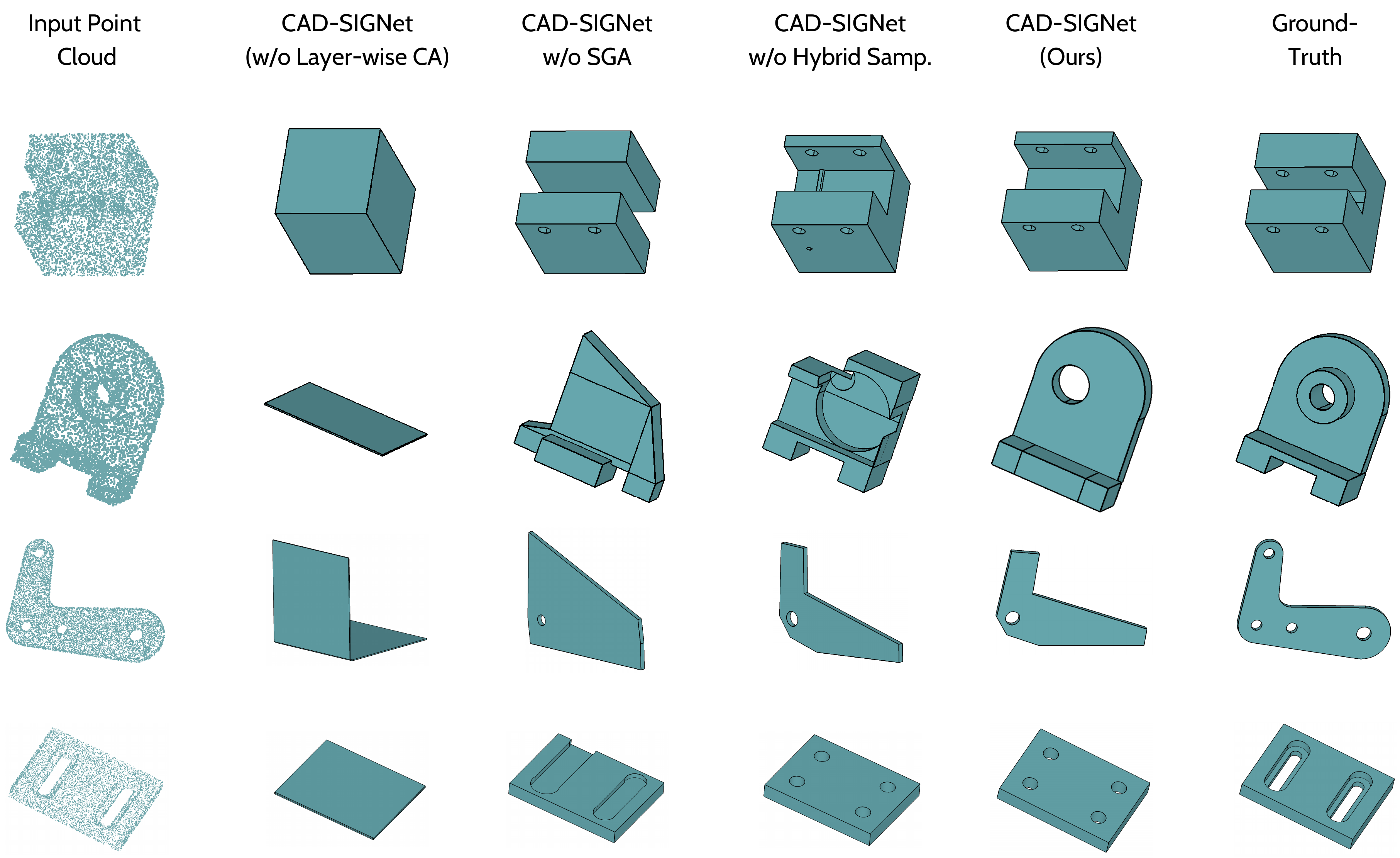}
    \caption{Reconstructed CAD models from predicted CAD sequences on DeepCAD~\cite{Wu_2021_ICCV} dataset for ablated versions of CAD-SIGNet.} 
    \label{fig:supp_ablation}
\end{figure*}

\begin{figure*}[ht]
    \centering
    \includegraphics[width=\linewidth]{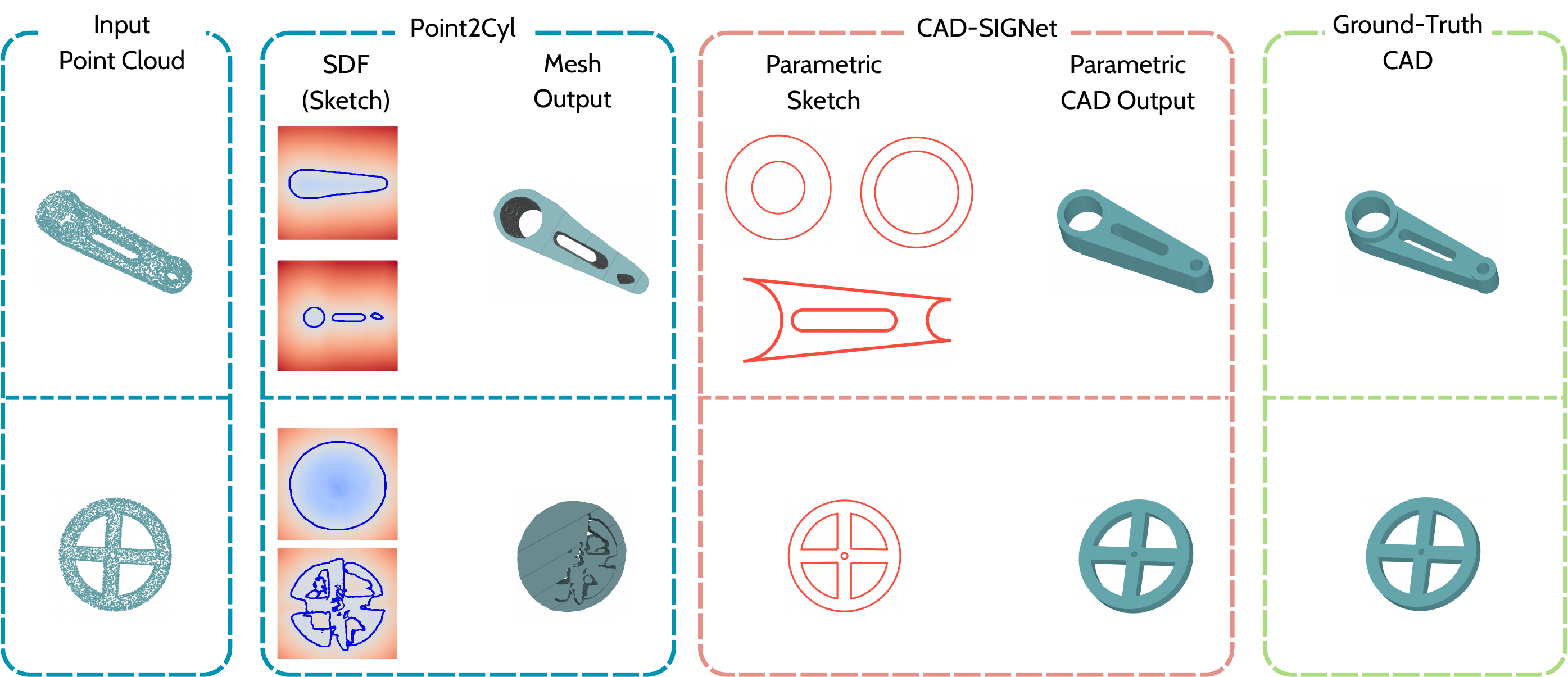}
    \caption{Qualitative results from Point2Cyl~\cite{uy2022point2cyl} and CAD-SIGNet. The output sketches and 3D reconstructions are provided here. The output from Poin2Cyl is not parametric while CAD-SIGNet outputs parametric CAD model.} 
    \label{fig:point2cyl}
\end{figure*}

\clearpage
\clearpage

\end{document}